\documentclass[runningheads]{llncs}
\usepackage{graphicx}
\usepackage{amsmath,amssymb} %
\usepackage{color}

\usepackage{algpseudocode}
\usepackage{algorithm}
\usepackage{pifont}
\usepackage{multirow}

\usepackage{booktabs}

\usepackage{caption}
\usepackage{subcaption}

\begin{document}
\title{AirBirds: A Large-scale Challenging Dataset for Bird Strike Prevention in Real-world Airports}
\titlerunning{AirBirds}
\author{Hongyu Sun \and
Yongcai Wang \and
Xudong Cai \and
Peng Wang \and
Zhe Huang \and \\
Deying Li \and
Yu Shao \and
Shuo Wang
}
\authorrunning{H. Sun et al.}
\institute{Renmin University of China, Beijing 100872, China\\
\email{\{sunhongyu,ycw,xudongcai,peng.wang,huangzhe21,\\deyingli,sy492019,shuowang18\}@ruc.edu.cn}}
\maketitle              %
\begin{abstract}
One fundamental limitation to the research of bird strike prevention is the lack of a large-scale dataset taken directly from real-world airports. Existing relevant datasets are either small in size or not dedicated for this purpose. To advance the research and practical solutions for bird strike prevention, in this paper, we present a large-scale challenging dataset AirBirds that consists of 118,312 time-series images, where a total of 409,967 bounding boxes of flying birds are manually, carefully annotated. The average size of all annotated instances is smaller than 10
pixels in 1920x1080 images. Images in the dataset are captured over 4 seasons of a whole year by a network of cameras deployed at a real-world airport, covering diverse bird species, lighting conditions and 13 meteorological scenarios. To the best of our knowledge, it is the first large-scale image dataset that directly collects flying birds in real-world airports for bird strike prevention. This dataset is publicly available at \url{https://airbirdsdata.github.io/}.

\keywords{Large-scale Dataset \and Bird Detection in Airport \and Bird Strike Prevention.}
\end{abstract}
\section{Introduction}
\begin{figure}[t]
  \centering
  \begin{subfigure}{0.98\linewidth}
    \includegraphics[width=\linewidth]{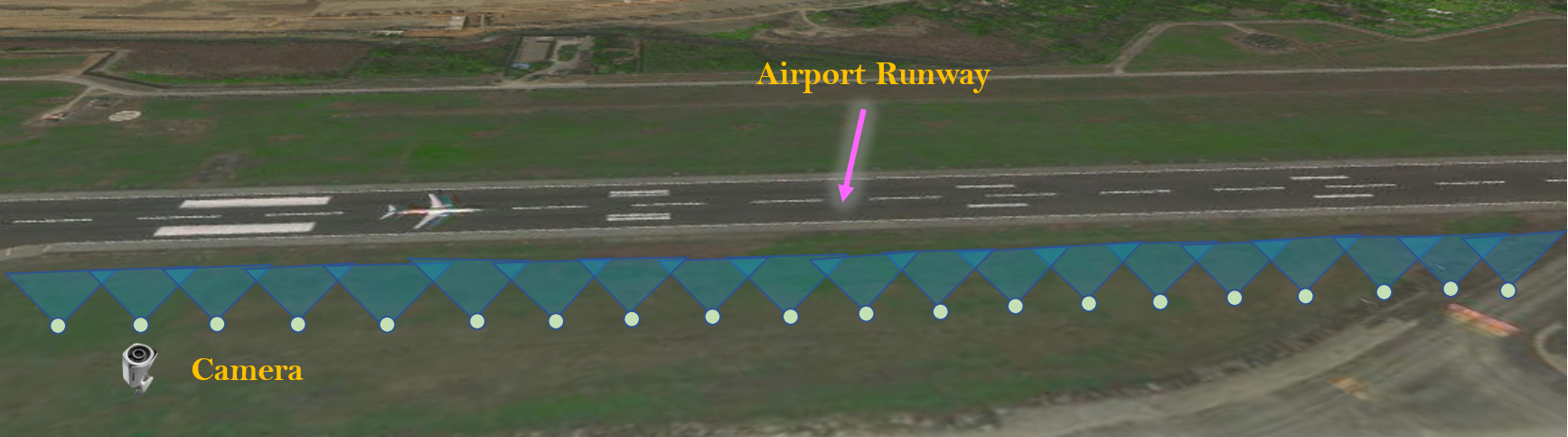}
    \caption{camera deployment alongside a runway in a real-world airport}
    \label{fig:camera_deployment}
  \end{subfigure}
  \hfill
  \begin{subfigure}{0.63\linewidth}
    \includegraphics[width=\linewidth]{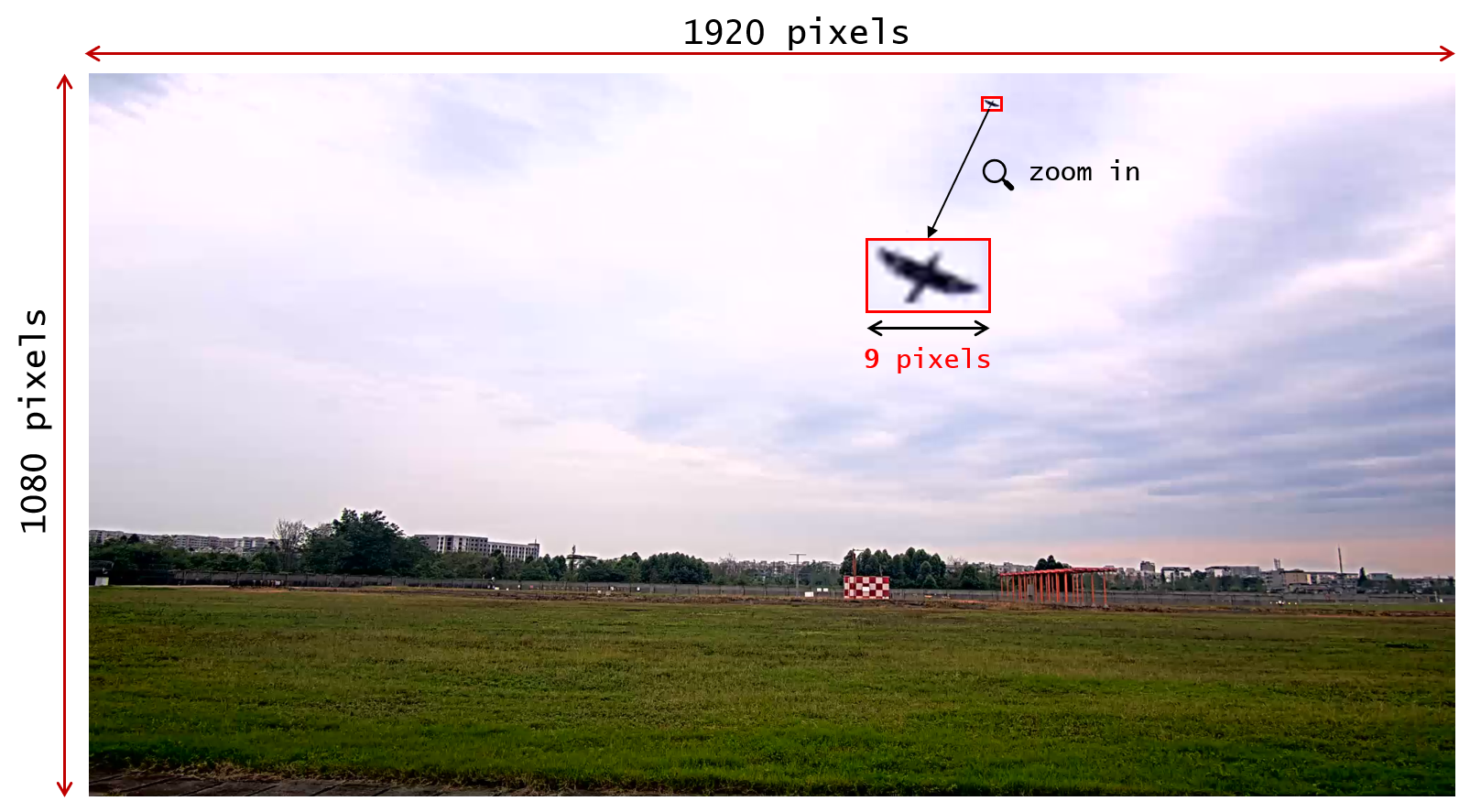}
    \caption{AirBirds}
    \label{fig:airbirds_examples}
  \end{subfigure}
  \hfill
  \begin{subfigure}{0.35\linewidth}
     \includegraphics[width=\linewidth]{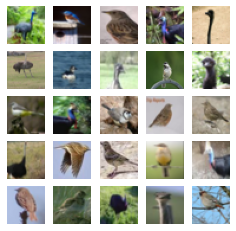}
     \caption{Relevant Datasets}
     \label{fig:other_dataset_examples}
  \end{subfigure}
  \caption{(a) The deployment of a network of cameras in a real-world airport (b) A bird example in AirBirds (c) Examples of birds in  CUB\cite{WelinderEtal2010,WahCUB_200_2011}, Birdsnap\cite{berg-birdsnap-cvpr2014}, NABirds\cite{Horn_2015_CVPR} and CIFAR10\cite{Krizhevsky09learning}.}
  \label{fig:overview}
\end{figure}

Bird strike accidents cause not only financial debts but also human
casualties. According to Federal Aviation Administration (FAA) \footnote{\url{https://www.faa.gov/airports/airport\_safety/wildlife/faq/}}, 
from 1990 to 2019, 
there have been more than 220 thousand wildlife strikes with civil 
aircraft in USA alone and 97\% of all strikes involve birds.
An estimated economic loss could be as high as \$500 million per year. Furthermore,
more than 200 human fatalities and 300 injuries attributed to bird strikes.
Bird strikes happen most near or at airports during takeoff, landing and associated phrases.
About 61\% of bird strikes with civil aircraft occur during landing phases of flight (descent, approach and landing roll). 36\% 
occur during takeoff run and climb\footnote{\url{https://en.wikipedia.org/wiki/Bird\_strike}}.
It is the airspace that the airport should be responsible for so that
the prevention of bird strikes is one of the most significant safety concerns.
Although various systems are designed for preventing bird strikes, accidents keep occurring
with increasing commercial activities and flights.
Improving the performances of bird strike prevention systems remains a research challenge.
One fundamental limitation to the performances is the lack of large-scale data collected at real-world airports.
On the one hand, 
real-world airports have strict rules on security and privacy 
regarding camera system deployment.
On the other hand, it is inevitably expensive to develop a large-scale dataset that involves a series of time-consuming
and laborious tasks.

Existing relevant datasets are either small in size or not dedicated for bird strike prevention.
The wildlife strike database created by FAA provides valuable information,
while each record in this database only contains a few fields in text form, such as date and time, aircraft and airport information, 
environment conditions, lacking informative pictures and videos.
The relevant dataset developed by Yoshihashi $et~al$ aims at preventing birds from hitting the blades of turbines in a wind 
farm~\cite{Yoshihashi15bird},
rather than in real-world airports, and its size is less than one seventh of ours.
Well-known datasets like ImageNet~\cite{deng09imagenet}, COCO~\cite{lin14coco}, VOC~\cite{Everingham10voc}, 
CIFAR~\cite{Krizhevsky09learning} collects millions of common 
objects and animals, including birds, but
they are developed for the research of general image recognition, object detection
and segmentation.
Another branch of datasets, such as
CUB series~\cite{WelinderEtal2010,WahCUB_200_2011}, Birdsnap~\cite{berg-birdsnap-cvpr2014} and NABirds~\cite{Horn_2015_CVPR} 
containing hundreds of bird species, focus on fine-grained categorization and part localization.
And the size of these datasets is less than 50\% of ours.
One of the most significant differences between the above-mentioned datasets and ours is that birds in previous datasets are carefully selected and tailored, 
which means they are often centered in the image, occupy the main part of an image and have clear outlines, referring to Fig.~\ref{fig:other_dataset_examples}. 

However, it is unlikely that birds in the images captured in real-world airports have these idealized characteristics. 
The deployment of a network of cameras around a runway in a real-world airport is shown in Fig.~\ref{fig:camera_deployment}.
Each camera 
is responsible for monitoring an area of hundreds of meters so that flying birds that appear are tiny in size even in a high-resolution 
image. For example, in our dataset, the average size of all annotated birds is smaller than 10 pixels in the 1920$\times$1080 images, taking up only $\sim$0.5\% of 
the image width, shown in Fig.~\ref{fig:airbirds_examples}.

To advance the research and practical solutions for bird strike prevention, 
we collaborate with a real-world airport for two years and finally present AirBirds, a large-scale challenging dataset consisting of 118,312 time-series 
images with 1920$\times$1080
resolution and 409,967 bounding box annotations
of flying birds. 
The images are extracted from videos recorded by a network of cameras over one year, from September 2020 to August 2021, thus 
cover various bird species in different seasons.
Diverse scenarios are also included in AirBirds, $e.g.$, changing lighting and 13 meteorological conditions. 
Planning, deployment and joint commissioning of the monitoring system last for one year. 
The data collection process takes another whole year and 
subsequent cleaning, labeling, sorting and experimental analysis consume parallel 12 months. 
To the best of our knowledge, AirBirds is the first large-scale challenging image dataset that collects flying birds in real airports for bird strike prevention.
The core contributions of this paper are summarized as follows.
\begin{itemize}
  \item A large-scale dataset, namely AirBirds, that consists of 118,312 time-series images with 1920$\times$1080 resolution containing flying birds in real-world airports
  is publicly presented, where there exist 409,967 instances with carefully manual bounding box annotations.
  The dataset covers various kinds of birds in 4 different seasons and diverse scenarios that
  include day and night, 13 meteorological and lighting conditions, $e.g.$, overcast, sunny, cloudy, rainy, windy, haze, etc. 
  \item 
  To reflect significant differences with other relevant datasets, we make comprehensive statistics on AirBirds and compare it with relevant datasets. 
  There are three appealing features. (i) The images in AirBirds are dedicatedly taken from a real-world airport, which provide rare first-hand sources
  for the research of bird strike prevention. (ii) Abundant bird instances in different seasons and changing scenarios are also covered by AirBirds as the data collection 
  spans a full year. 
  (iii) The distribution of AirBirds is distinctive with existing datasets since 88\% of instances are smaller than 10 pixels, and the remaining 12\%
  are more than 10 and less than 50 pixels in 1920$\times$1080 images. 
  \item 
  To understand the difficulty of AirBirds, a wide range of strong baselines are evaluated on this dataset for bird discovering. Specifically,
  16 detectors are trained $from~ scratch$ based on AirBirds with careful configurations and parameter optimization. The consistently unsatisfactory results  
  reveal the non-trivial challenges of bird discovering and bird strike prevention in real-world airports, which deserve further investigation. 
\end{itemize}
As far as we know, bird strike prevention remains a open research problem 
since it is not well solved by existing technologies. 
We believe AirBirds will benefit the researchers, facilitate the research 
field and push the boundary of practical solutions in real-world airports. 

\section{Related Work}
\label{sec:blind}

In this section, we review the datasets 
that are either closely relevant to bird strike prevention or 
contain transferrable information to this topic. 

\begin{table*}[t]
	\small
   \centering
   \caption{Comparisons of AirBirds and relevant datasets. Density is the average instances in each image. Duration refers to the period of data collection.}
   \begin{tabular}{l r r r r r r}
     \toprule
     \textbf{Dataset} & \textbf{Format} & \#\textbf{Images} & \textbf{Resolution} & \#\textbf{Instances} & \textbf{Density} & \textbf{Duration}\\
     \midrule
     FAA Database & text & - & - &  227,005 & - & 30 years \\
     CUB-200-2010~\cite{WelinderEtal2010} & image & 6,033 & $\sim$500$\times$300 & 6,033 & 1.00 & - \\
     CUB-200-2011~\cite{WahCUB_200_2011} & image & 11,788 &$\sim$500$\times$300 &  11,788 & 1.00 & - \\
     Birdsnap~\cite{berg-birdsnap-cvpr2014} & image & 49,829 & various &  $\sim$49,829 & 1.00 & - \\
     NABirds~\cite{Horn_2015_CVPR} & image & 48,562 & various &  $\sim$48,562 & 1.00 & - \\
     Wind Farm~\cite{Yoshihashi15bird} & image & 16,200 & $\sim$5616$\times$3744 & 32,000 & 1.97 & 3 days \\
     VB100~\cite{ge2016temporal} & video & - & $\sim$848$\times$464 & 1,416 & - & - \\
     \midrule
     \textbf{AirBirds} & image & 118,312 & 1920$\times$1080 & 409,967 & 3.47 & 1 year \\ 
     \bottomrule
   \end{tabular}
   \label{tab:comparison_relevant_dataset}
\end{table*}   
   
\textbf{FAA Wildlife Strike Database}.\quad
One of the most relevant datasets is the Wildlife Strike Database\footnote{\url{https://wildlife.faa.gov}} maintained by FAA. This database contains more than 220K records of 
reported wildlife strikes since 1990
and 97\% of strikes attribute to birds.  
The detailed descriptions for each incident can be divided into the following parts: bird species, 
date and time, airport information, 
aircraft information, environment conditions, etc. 
An obvious limitation is the contents in this database are mainly in text form, lacking  
informative pictures and videos.

\textbf{Bird Dataset of a Wind Farm}. 
Yoshihashi $et~al$ develop this dataset for preventing birds striking the 
blades of the turbines in a wind farm~\cite{Yoshihashi15bird}. 
32,000 birds and 4,900 non-birds are annotated in total to conduct experiments of a two-class categorization. 
It is similar to us that the ratio of bird size and the image size is extremely small. 
However, compared to AirBirds' data collection process spanning a whole year, 
this dataset collects images only for 3 days so that the number of samples and scenarios are 
much less than those of AirBirds.

\textbf{Bird Datasets with Multiple Species}.
Bird species probably provide valuable information for bird strike prevention. Another branch of the relevant datasets, 
such as CUB series~\cite{WelinderEtal2010,WahCUB_200_2011}, Birdsnap~\cite{berg-birdsnap-cvpr2014},
NABirds~\cite{Horn_2015_CVPR} and VB100~\cite{ge2016temporal},
focuses on fine-grained categorization of bird species. 
Images in these datasets are mainly collected from public sources, $e.g.$, Flickr~\footnote{\url{https://www.flickr.com/search/?text=bird}}, or by professionals.
One of the most significant differences between the datasets in this branch and AirBirds is that birds in these datasets are carefully tailored, 
which means they are often centered in the image, occupy the main part and have clear outlines. 
However, it is unlikely for birds captured in real-world airports to have these wonderful characteristics.
Moreover, bounding box annotations are absent in some of them, $e.g.$, VB100~\cite{ge2016temporal},
thus they are not suitable for the research of tiny bird detection.

\textbf{Well-Known Datasets Containing Birds}. Commonly used datasets in computer vision are also relevant
as the \verb|bird| belongs to
one of the predefined categories in those datasets and there exist numbers of samples, 
such as ImageNet~\cite{deng09imagenet}, COCO~\cite{lin14coco}, VOC~\cite{Everingham10voc}, 
CIFAR~\cite{Krizhevsky09learning}. 
However, the above-mentioned datasets are dedicatedly designed for the research of general image classification, object detection and 
segmentation, not for bird strike prevention.
And their data distributions differ from AirBirds,
thus limited information can be transferred to this task.

The comparisons with related work are summarized in Tab.~\ref{tab:comparison_relevant_dataset}. 
AirBirds offers
the most instances, the longest duration and the richest scenarios in image form.

\begin{figure}[t]
  \begin{center}
    \includegraphics[width=0.75\linewidth]{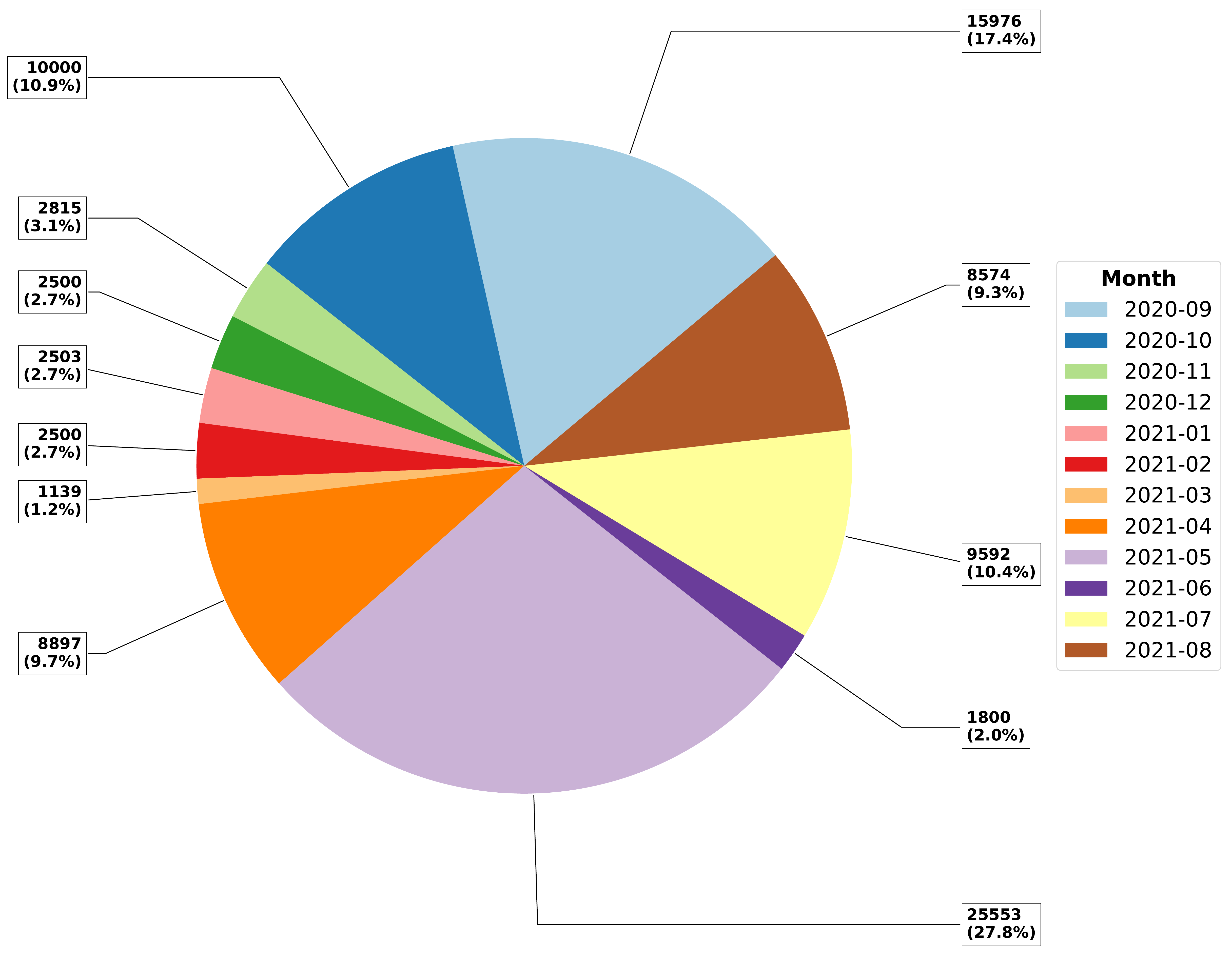}
  \end{center}
  \caption{The number of images per month in AirBirds.}
  \label{fig:number_of_images_per_month}
\end{figure}

\section{AirBirds Construction}
\label{sec:airbirds_construction}

This section describes the process of constructing the AirBirds dataset, including raw data collection, 
subsequent cleaning, annotation, splits and sorting to complete it.

\subsection{Collection}
To cover diverse scenarios and prepare adequate raw data, we decide to record 
in a real-world airport (Shuangliu International Airport, Sichuan Province, China) over 4 seasons of a whole year. The process of data collection starts from September 
2020 and ends in August 2021. 

Considering frequent takeoffs and landings, airport runways and their surroundings are major monitoring areas.
We deployed a network of high-resolution cameras along the runways, as Fig.~\ref{fig:camera_deployment} shows.
All deployed cameras use identical configurations.
The camera brand is AXIS Q1798-LE\footnote{\url{https://www.axis.com/products/axis-q1798-le}}, 
recording 1920$\times$1080 images at a frame rate of 25.
Due to the vast volume of raw data but a limited number of disks, it is infeasible to save all videos.
We split into two parallel groups, one group for data collection and the other for data processing,
so disk spaces can be recycled once the second group finishes data processing.

\subsection{Preprocessing}

\begin{figure}[t]
  \begin{center}
     \includegraphics[width=0.7\linewidth]{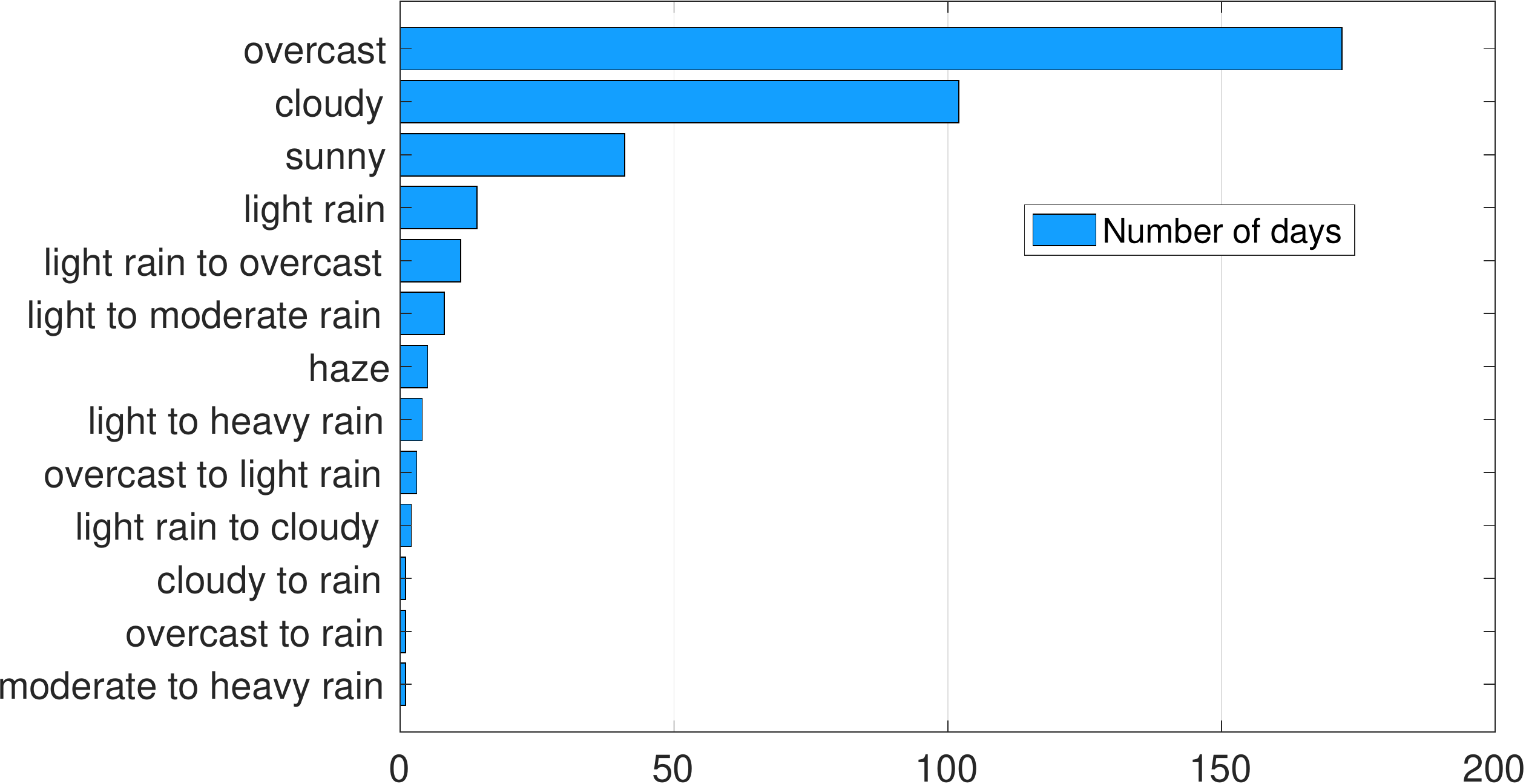}
  \end{center}
     \caption{The number of days of different weather in AirBirds.}
  \label{fig:weather_and_count}
\end{figure}

This step aims to process raw videos month by month and save 1920$\times$1080 images in chronological order. 
25 frames per second in raw videos lead to numerous redundant images. 
To avoid dense distribution of similar scenarios, a suitable sampling strategy is required.
One crucial observation is that the video clips where flying birds appear are very sparse compared to other clips.
Hence, at first, we manually locate all clips where there exist birds, then sample one every 5 continuous frames in previously 
selected clips instead of all of them, resulting in an average of 300+ images per day, $\sim$10000 images per month, 118,312 in total. 
The number of images per month is shown in Fig.~\ref{fig:number_of_images_per_month}
and 13 meteorological conditions and the corresponding number of days are depicted in Fig.~\ref{fig:weather_and_count}.

\subsection{Annotation}

\label{subsec:annotation}
To ensure quality and minimize costs,
we divide the labeling process into three rounds. The first round that generates initial bounding box annotations
for birds in the images is done by machines. The second round refines previous annotations manually by a team of 
employed workers. It should be noted that the team does not have to discover birds from scratch. In the third round, 
we are responsible for verifying 
those manual annotations and requiring further improvements of low-quality instances.

\begin{algorithm}[t]
  \caption{The First Round of Annotations}\label{alg:detection}
  \hspace*{\algorithmicindent} \textbf{Input}: $\mathcal{S} =  \{I_1, I_2, \dots, I_n\}$, an image sequence, where $n$ is the sequence length.
  Constants $min$ and $max$\\
  \hspace*{\algorithmicindent} \textbf{Output}: $\mathcal{B} = \{\textbf{b}_1, \textbf{b}_2, \dots, \textbf{b}_n\}$, the bounding
  boxes set for birds in $\mathcal{S}$, where $\textbf{b}_i = \{b^1_i, b^2_i, \dots, b^m_i\}$, $m$ is the number of detected
  birds in image $I_i$
  \begin{algorithmic}[1]
  \State $g \gets $ \verb|imRead|($I_1$, 0) \Comment{read image in gray mode}
  \State $b$ $\gets$ \verb|gaussBlur|$(g)$ \Comment{denoise the image}
  \State $\mathcal{S} \gets \mathcal{S}~\backslash~I_1$  \Comment{remove $I_1$ from $\mathcal{S}$}
  \State $\mathcal{B} \gets \emptyset$  \Comment{initialize $B$}
  \For{$I_i$ in $\mathcal{S}$}
    \State $g_i$ $\gets$ \verb|imRead|$(I_i, 0)$
    \State $c_i$ $\gets$ \verb|gaussBlur|$(g_i)$
    \State $d$ $\gets$ \verb|Diff|$(b$, $c_i)$  \Comment{compute differences}
    \State $d$ $\gets$ \verb|Thresh|$(d$, $min$, $max)$  \Comment{apply threshold}
    \State $d_i$ $\gets$ \verb|Dilate|$(d)$  \Comment{dilate areas further}
    \State $\textbf{c}_i$ $\gets$ \verb|findContours|$(d_i)$ \Comment{find candidates}
    \State $\textbf{b}_i$ $\gets$ \verb|Filter|$(\textbf{c}_i)$ \Comment{filter candidates}
    \State $\mathcal{B}$ $\gets$ $\mathcal{B} \cup \textbf{b}_i$  \Comment{insert annotations}
    \State $b \gets c_i$   \Comment{move background next}
  \EndFor
  \end{algorithmic}
\end{algorithm}

\begin{figure}[t]
  \begin{center}
     \includegraphics[width=0.97\linewidth]{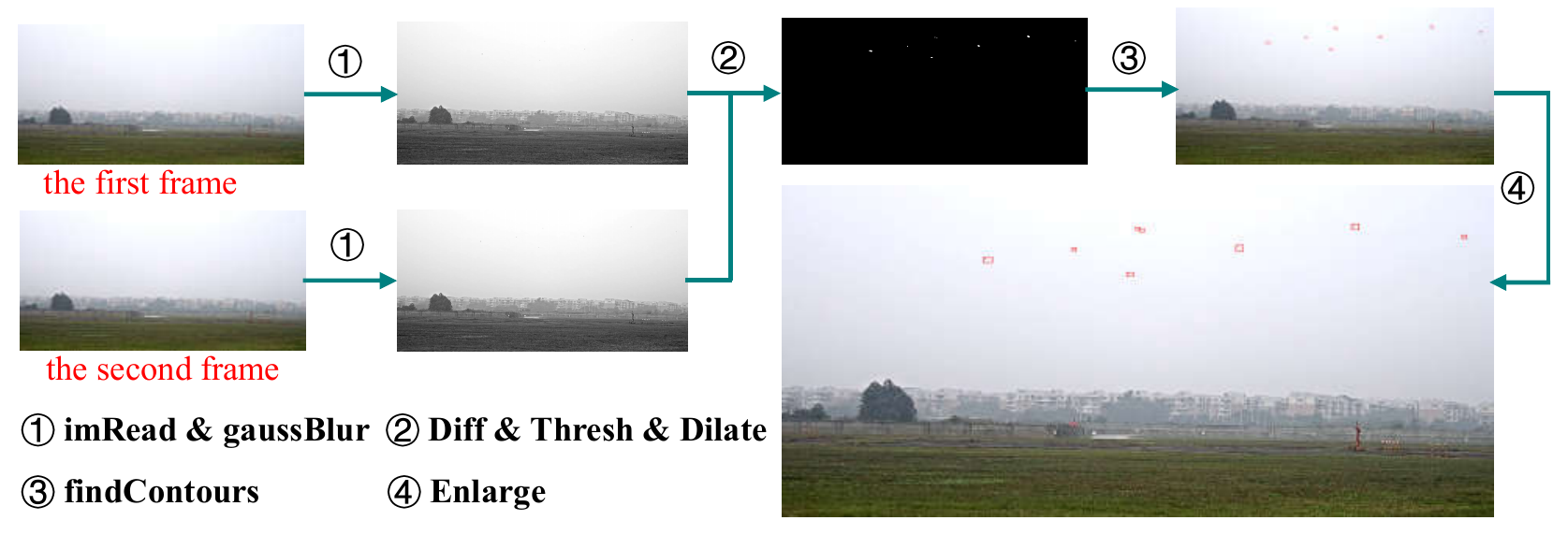}
  \end{center}
     \caption{Visualization of key steps in Alg.~\ref{alg:detection}.}
  \label{fig:first_round_annotation}
\end{figure}

It is not a simple task for humans to discover tiny birds from collected images with broad scenes.
In the first round, we develop an algorithm for generating initial annotations and run on a computer. 
The idea of this algorithm is related to background subtraction in image processing. In our context, 
cameras are fixed in real-world airports,
thus the background is static in the monitoring views. 
Since the images in each sequence are in chronological order, considering two consecutive frames, 
by computing the pixel differences between the first and second frame, the static part, namely the background, is removed while 
other moving targets, such as flying birds in the monitoring areas, are probably discovered.
Alg.~\ref{alg:detection} describes the detailed process.
Initially, we treat the first frame as background, convert it to gray mode, apply Gaussian blur\footnote{\url{https://en.wikipedia.org/wiki/Gaussian\_blur}} to this gray image, and denote the
 output image as $b$, then remove the first image from the input sequence $\mathcal{S}$.
The set of initial bounding box annotations $\mathcal{B}$ is empty.
Then we traverse the image $I_i$ in $\mathcal{S}$.
In the loop, $I_i$ is also converted to gray image $g_i$.
After that, Gaussian blur is applied to $g_i$ to generate a denoised
image $c_i$. 
Then we compute differences between $b$ and $c_i$, resulting in $d$.
Fourth, regions in $d$ whose pixel values are in the range of [$min$, $max$] are considered as areas of interest, $e.g.$, if the pixel
differences of the same area in those 2 consecutive frames are more 
than 30, there probably are moving targets in this area.
The dilation operation is applied to those areas to expand contours for finding possible moving objects $\textbf{c}_i$, 
including flying birds. After that, heuristic rules are used to filter candidates
according to the object size, $e.g.$, big targets like airplanes, working vehicles, workers, are removed,
resulting in $\textbf{b}_i$.
Then $\textbf{b}_i$ is inserted into $\mathcal{B}$.
Finally, we set background $b$ as $c_i$ and move forward.
The key steps of this algorithm are visualized in Fig.~\ref{fig:first_round_annotation}. 

Refinement is required since previously discovered moving objects are not necessarily birds.
In the second round, we cooperate with a team of workers to accomplish the task. According to the predefined instructions, 
every single image should be zoomed in to 250+\% to check the initial annotations in detail and
the team mainly handles 3 types of 
issues that arose in the first round
(i) add missed annotations, (ii) delete false-positive annotations, (iii) update inaccurate 
annotations. 
In the third round, we go through the annotations refined by the team, requiring further improvements where inappropriate.

\subsection{Splits}

To facilitate further explorations of bird strike prevention based on this dataset, it is necessary
to split AirBirds into training and test set.

We need to pay attention to three key aspects when splitting the dataset. 
First, we should keep a proper ratio between the size of the training and the test set. Second, 
it is essential to ensure training and test sets have a similar distribution.
Third, considering the characteristic of chronological order, we should 
put a complete sequence into either the training or the test set rather than split it into
different sets. 

At last, we divide 98,312 images into the training set and keep the remaining 20,000 images in the test set, a nearly 5:1 ratio. 
All images and labels are publicly available, but excluding the labels in the test set.
The validation set is not explicitly distinguished as the primary evaluation should take place on the test set,
and users can customize the ratio between training and validation set individually.
We are actively building an evaluation server and the labels in the test will be kept there.

In addition, the images in AirBirds can also be divided into 13 groups according to 13 kinds of scenarios shown in 
Fig.~\ref{fig:weather_and_count}. This division is easy to achieve since each image is recorded on a specific day and 
each day corresponds to one type of meteorological condition, according to the official weather report. 
Based on the division, we can evaluate the difficulty of bird discovering in different
scenarios in real-world airports. 

\section{Experiments}
\label{sec:exp}

In this section, a series of comprehensive statistics and experiments based on AirBirds 
are presented. 
First, we investigate the data distribution in AirBirds and compare with relevant datasets
to reflect their significant differences. 
Second, a wide range of SOTA detectors are evaluated on the developed dataset
for bird discovering and the results are analyzed in detail to understand the non-trivial challenges of bird strike prevention. 
Third, the effectiveness of Alg.~\ref{alg:detection} is evaluated since it plays
an important role in the first round of annotations when constructing AirBirds. 

\begin{figure}
  \centering
  \begin{subfigure}{0.49\linewidth}
    \includegraphics[width=\linewidth]{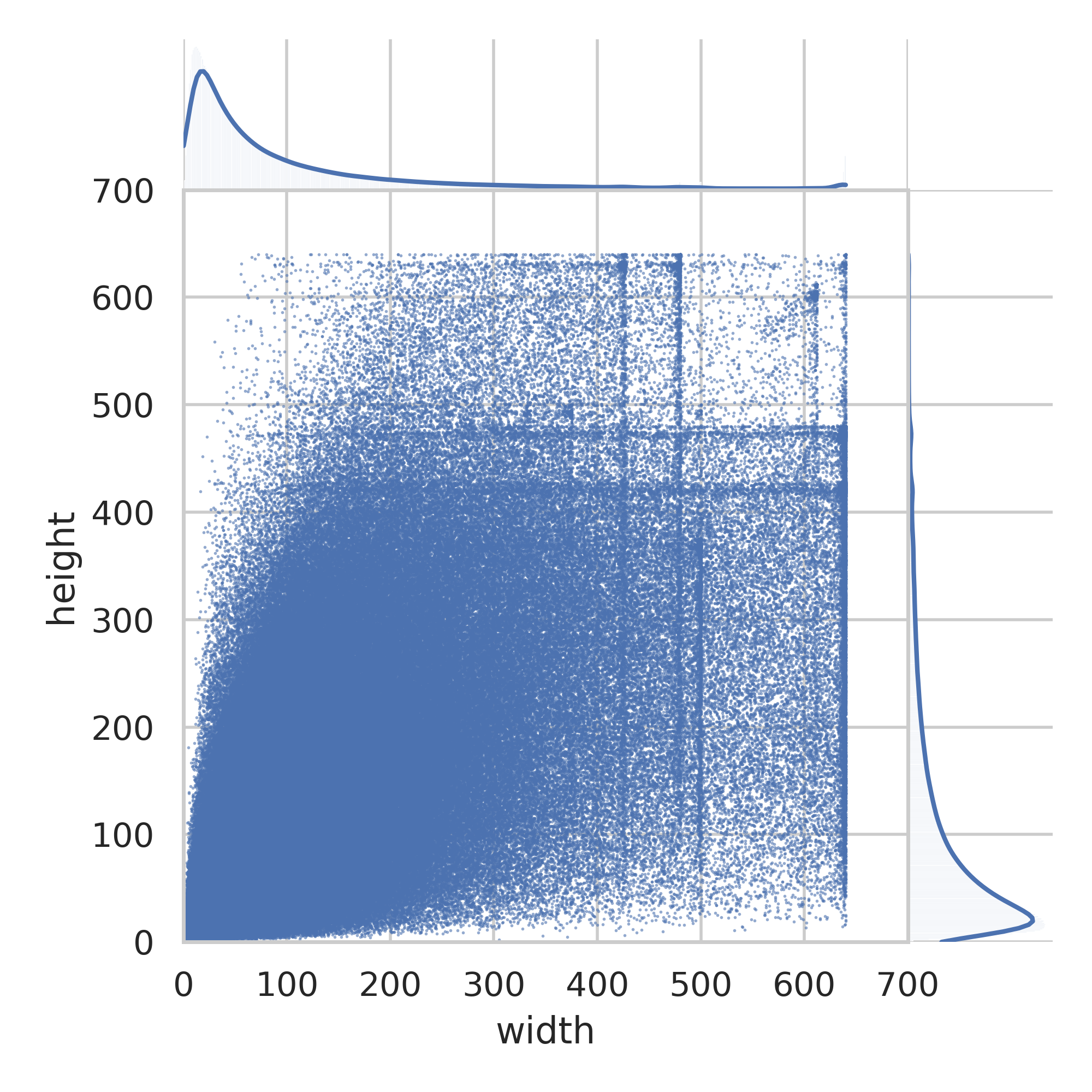}
    \caption{COCO}
    \label{fig:coco_bwd}
  \end{subfigure}
  \hfill
  \begin{subfigure}{0.49\linewidth}
     \includegraphics[width=\linewidth]{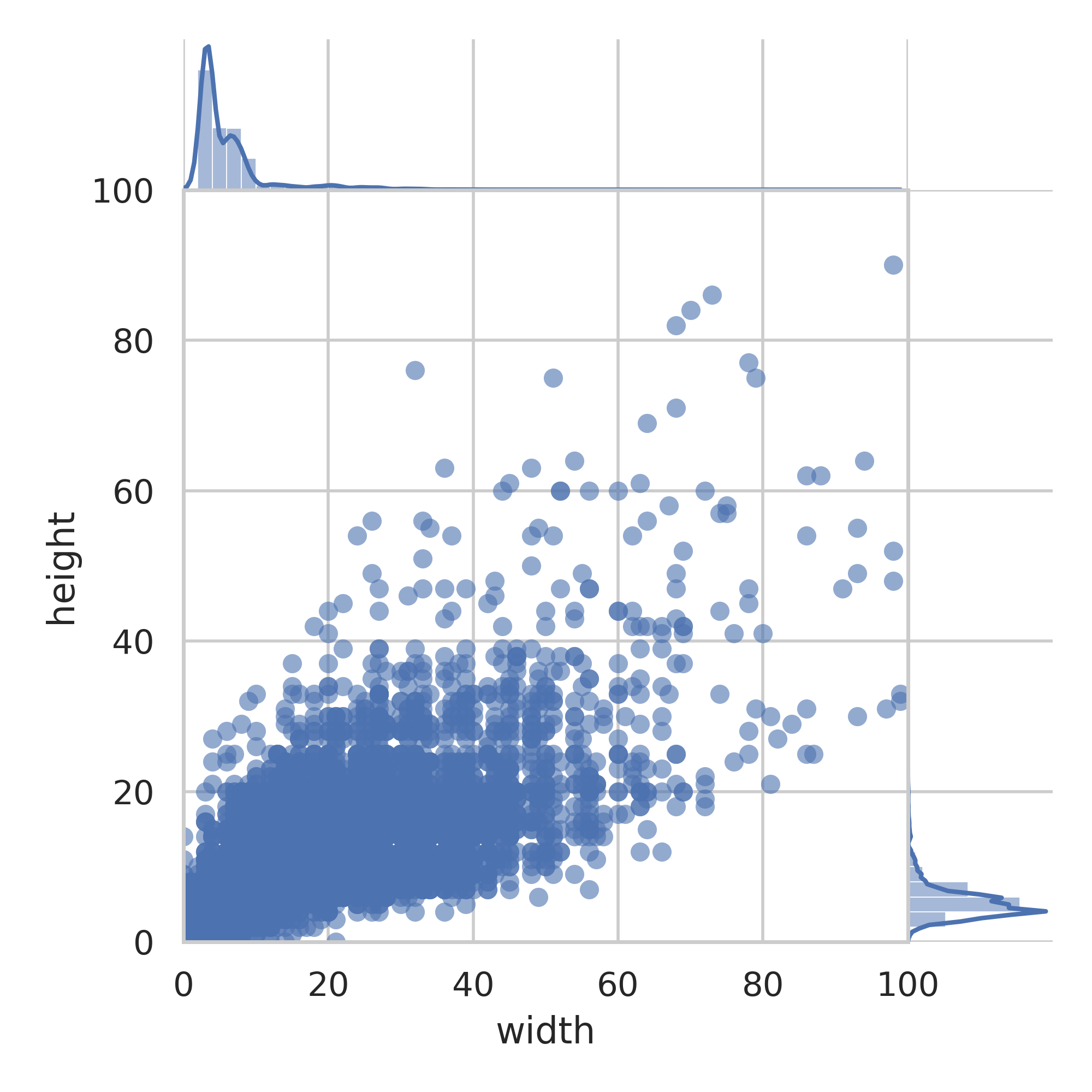}
    \caption{AirBirds}
    \label{fig:airbirds_bwd}
  \end{subfigure}
  \hfill
  \caption{Distributions of width and height of annotated instances in COCO and AirBirds.}
  \label{fig:box_width_height}
\end{figure}

\subsection{Distribution}
\begin{figure}[ht]
  \begin{center}
     \includegraphics[width=0.75\linewidth]{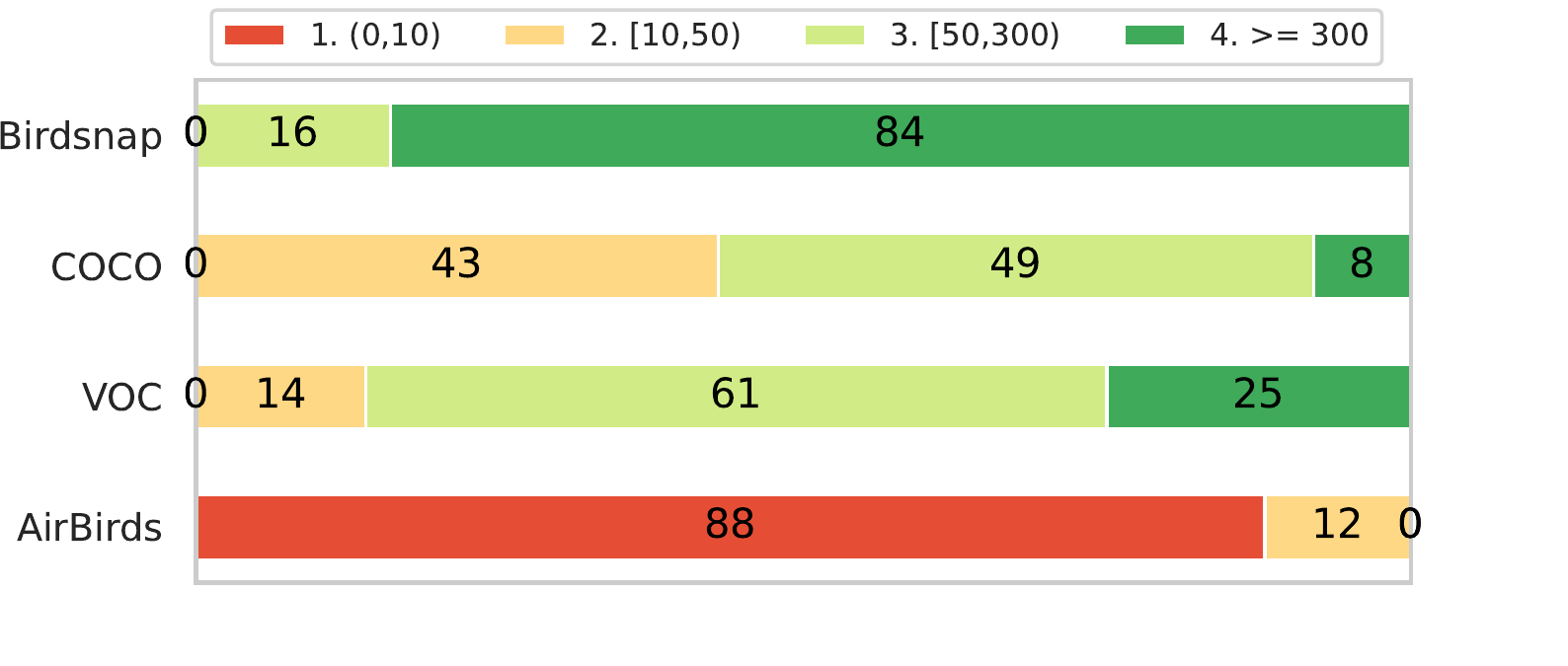}
  \end{center}
     \caption{Comparisons of the ratio of the number of objects with different sizes in the datasets Birdsnap, COCO, VOC and AirBirds. 
     The numbers in each bar are in \%. 
     Here the object size in pixel level is divided
     into 4 intervals: (0,10), [10,50), [50, 300) and [300,$+\infty$).}
  \label{fig:object_ratio_comparison}
\end{figure}

In this subsection, we investigate the distribution of AirBirds and compare with relevant datasets.
Figure~\ref{fig:box_width_height} shows the distibution of width and height of bounding box in different datasets.
Obviously, objects in AirBirds have much smaller sizes. Further, Fig.~\ref{fig:object_ratio_comparison} depicts the 
proportion of objects with various sizes in relevant datasets.
88\% of all instances in AirBirds are smaller than 10 pixels and
the rest 12\% are mainly in the interval [10, 50). 
Therefore, data distribution in real-world airports is significantly different from that in web-crawled and tailor-made datasets.

\begin{table*}[t]
  \centering
  \caption{Comparisons of various kinds of object detectors on AirBirds test set.
  The column AP$_l$ has been removed as there
  are few large objects in AirBirds and the corresponding scores are all 0. 
  EffiDet: EfficientDet-D2, Faster: Faster RCNN, Cascade: Cascade RCNN, 
  Deform: Deformable DETR. These abbreviations have the same meaning in the following figure or table. 
  }
  \label{tab:detection_results_on_airbirds}
  \begin{tabular}{l | l | l | r r r r r r}
          \toprule
          Method & Type & Backone & AP & AP@50 & AP@75 & AP$_s$ & AP$_m$ & FPS\\
          \midrule
          FCOS\cite{tian2019fcos} &  & ResNet50\cite{he16resnet} & 0.3 & 1.3 & 0.0 & 0.3 & 0.2 & 18.5\\ %
          EffiDet\cite{tan20effidet} & \multirow{2}{*}{one-stage} & EffiNet-B2\cite{tan2019efficientnet} & 0.6 & 1.0 & 1.0 & 0.6 & 4.3 & 4.88 \\
          YOLOv3\cite{redmon2018yolov3} & & DarkNet53\cite{darknet13} & 5.8 & 24.1 & 1.4 & 5.8 & 6.8 & 19.5\\ %
          YOLOv5\cite{Ultralytics21yolov5} & & CSPNet\cite{Wang20CSPNet} &11.9 & 49.5 & - & - & - & 109.9 \\
          \midrule
          Faster\cite{ren17faster_rcnn} & \multirow{2}{*}{multi-stage} & \multirow{2}{*}{ResNet50\cite{he16resnet}} &7.1 & 26.9 & 1.3 & 7.1 & 0.2 & 16.0 \\  %
          Cascade\cite{Cai_2018_CVPR} & & & 6.8 & 24.0 & 1.8 & 6.8 & 1.8 & 13.4 \\
          \midrule
          DETR\cite{nicolas20detr} & \multirow{2}{*}{transformer} & \multirow{2}{*}{ResNet50\cite{he16resnet}} & 0.0 & 0.0 & 0.0 & 0.0 & 0.0 & 19.7\\
          Deform\cite{zhu2021deformable} & & & 0.4 & 2.2 & 0.0 & 0.4 & 1.0 & 11.6\\
          \midrule
          FPN\cite{Lin_2017_CVPR} & \multirow{2}{*}{FPN} & \multirow{2}{*}{RetinaNet}\cite{Lin_2017_ICCV} & 2.9 & 12.5 & 0.3 & 2.9 & 8.0 & 21.3\\ %
          NASFPN\cite{Ghiasi_2019_CVPR} & & & 3.0 & 12.5 & 0.4 & 2.9 & 15.8 & 25.0\\  %
          \midrule
          RepPoints\cite{yang2019reppoints} & \multirow{3}{*}{anchor-free} & ResNet50\cite{zhou2019objects} & 4.8 & 22.6 & 0.3 & 4.9 & 0.0 & 37.1\\ %
          CornerNet\cite{law2018cornernet} & & HourglassNet & 4.5 & 19.5 & 0.6 & 5.0 & 2.5 & 5.5\\ %
          FreeAnchor\cite{zhang2019freeanchor} & & ResNet101\cite{he16resnet} & 6.6 & 26.5 & 1.1 & 6.7 & 9.0 & 53.5\\ %
          \midrule
          HRNet\cite{SunXLW19} & high-resolution & HRNet\cite{SunXLW19} & 8.9 & 33.0 & 1.3 & 9.0 & 0.3 & 21.7\\ %
          DCN\cite{dai2017deformable} & \multirow{2}{*}{deformable} & ResNet50\cite{he16resnet} & 9.7 & 34.6 & 1.8 & 9.8 & 2.4 & 41.9\\
          DCNv2\cite{zhu2018deformable} & & ResNet50\cite{he16resnet} & 4.2 & 17.5 & 0.5 & 4.6 & 0.0 & 14.2\\  %
          \bottomrule
  \end{tabular}
\end{table*}

\begin{table}[t]
  \centering
  \caption{Comparisons of Alg.~\ref{alg:detection} and YOLOv5 in terms of precision, recall and f1 score. Alg.~\ref{alg:detection} runs on a common
  computer and YOLOv5 is tested with a 2080Ti GPU.}
  \begin{tabular}{l l l c c}
          \toprule
          Method & Precision & Recall & F1 & FPS\\
          \midrule
          YOLOv5 & \textbf{68.10\%} & 55.50\% & 61.16\% & \textbf{109.89}\\
          Alg.~\ref{alg:detection} & 58.29\% & \textbf{95.91\%} & \textbf{72.51}\% & 67.44$\star$\\  
          \bottomrule
  \end{tabular}
  \label{tab:first_round_p_r_f1}
\end{table}
\subsection{Configurations}
A wide range of detectors are tested on AirBirds for bird discovering. Before reporting their performances, 
it is necessary to elaborate on the specific modifications we made to accommodate the AirBirds dataset and the detectors. 
Concretely, we customize the following settings.

\textbf{Models}. To avoid AirBirds preferring a certain type of detectors, various kinds of strong baselines are picked for evaluation, 
including one-stage, multi-stage, transformer-based, anchor-free, 
and other types of models, 
referring to Tab.~\ref{tab:detection_results_on_airbirds}.

\textbf{Devices}. 6 NVIDIA RTX 2080Ti GPUs are used during training and a single GPU device is used during test for all models.

\textbf{Data Format}. The format of annotations in AirBirds is consistent with YOLO~\cite{Redmon16yolo} style. Then we convert them to
COCO format when training models other than YOLOv5.

\textbf{Anchor Ratios and Scales}. 
We need to adapt the ratios and 
scales for anchor-based detectors to succeed in custom training because objects in AirBirds
have notable differences in size with that in the commonly used COCO dataset. 
The k-means clustering is applied to the labels of AirBirds, 
finally the ratios are set to [$\frac{8}{13}$, $\frac{9}{12}$, $\frac{11}{9}$] and the scales are set to 
[2$^0$, 2$^{\frac{1}{3}}$, 2$^{\frac{2}{3}}$].

\textbf{Learning Schedules}. All models are trained $from~ scratch$ with optimized settings, $e.g.$, training epochs, learning rate, optimizer,
batch size, etc. We summarize these settings in Section 2 in the supplementary material.

\textbf{Alg.}~\ref{alg:detection}. The thresholds of pixel differences in Alg.~\ref{alg:detection}, 
$min$ and $max$, are set to 25 and 255, respectively.

\begin{figure}[t]
  \begin{center}
     \includegraphics[width=0.7\linewidth]{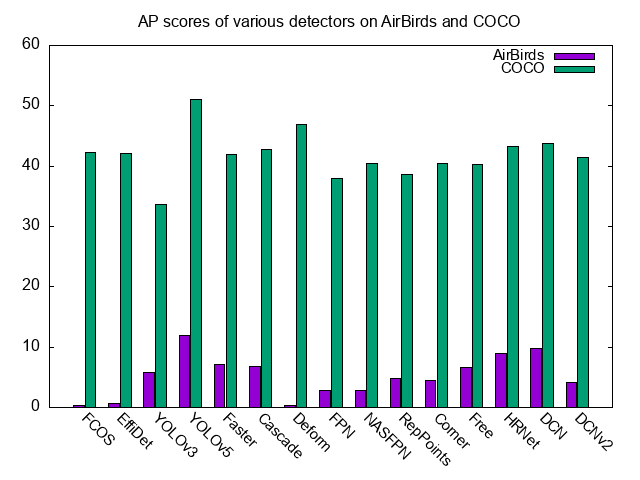}
  \end{center}
  \caption{Comparisons of the performances among representative detectors on AirBirds and COCO.}
  \label{fig:map_coco_airbird}
\end{figure}

\begin{figure}[!]
  \centering
  \begin{subfigure}{0.24\linewidth}
    \includegraphics[width=\linewidth]{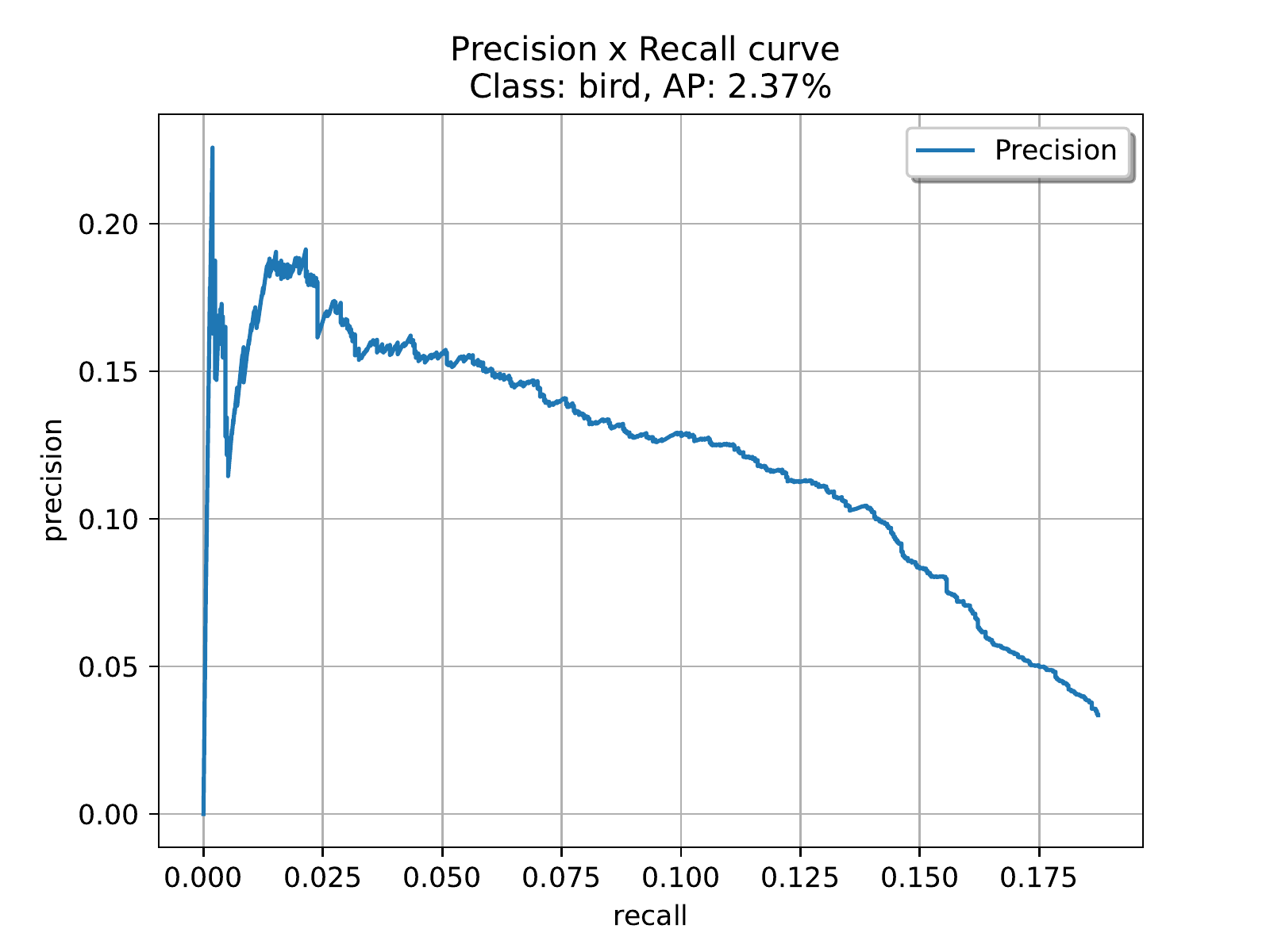}
    \caption{FCOS}
    \label{fig:voc_pr_curve_fcos}
  \end{subfigure}
  \begin{subfigure}{0.24\linewidth}
     \includegraphics[width=\linewidth]{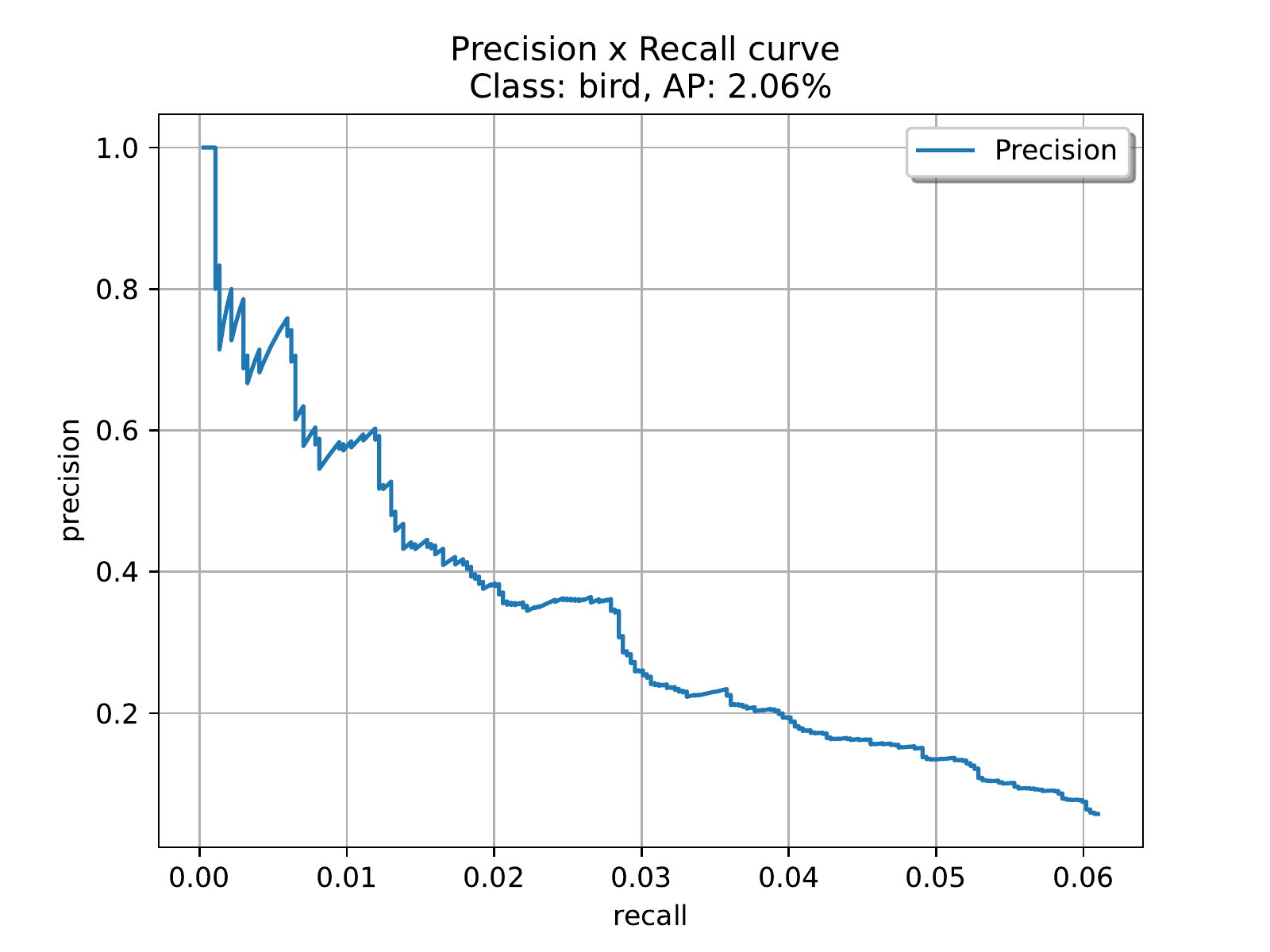}
     \caption{EffiDet}
     \label{fig:voc_pr_curve_effidet}
  \end{subfigure}   
  \begin{subfigure}{0.24\linewidth}
    \includegraphics[width=\linewidth]{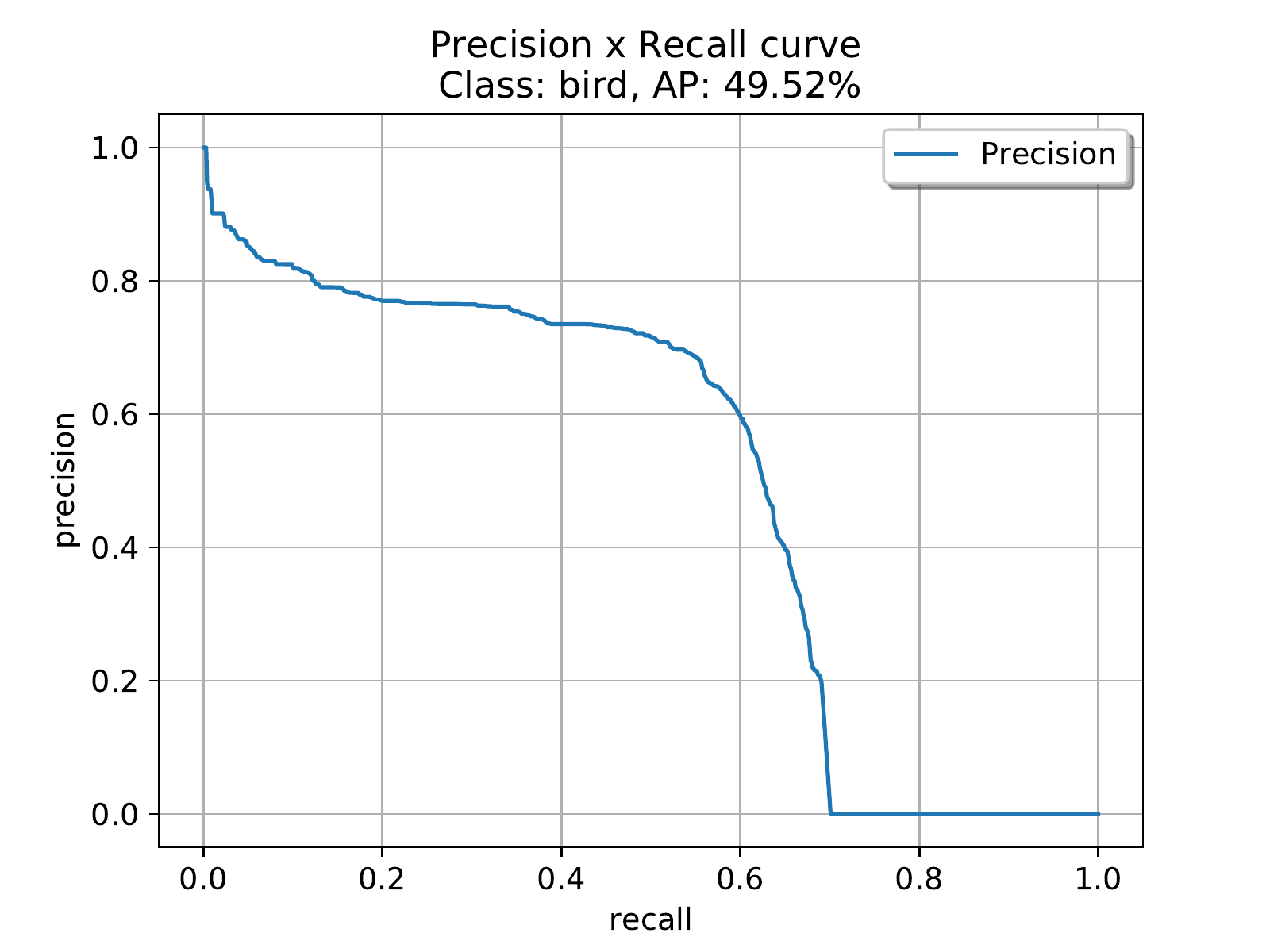}
    \caption{YOLOv5}
    \label{fig:voc_pr_curve_yolov5}
  \end{subfigure}
  \begin{subfigure}{0.24\linewidth}
    \includegraphics[width=\linewidth]{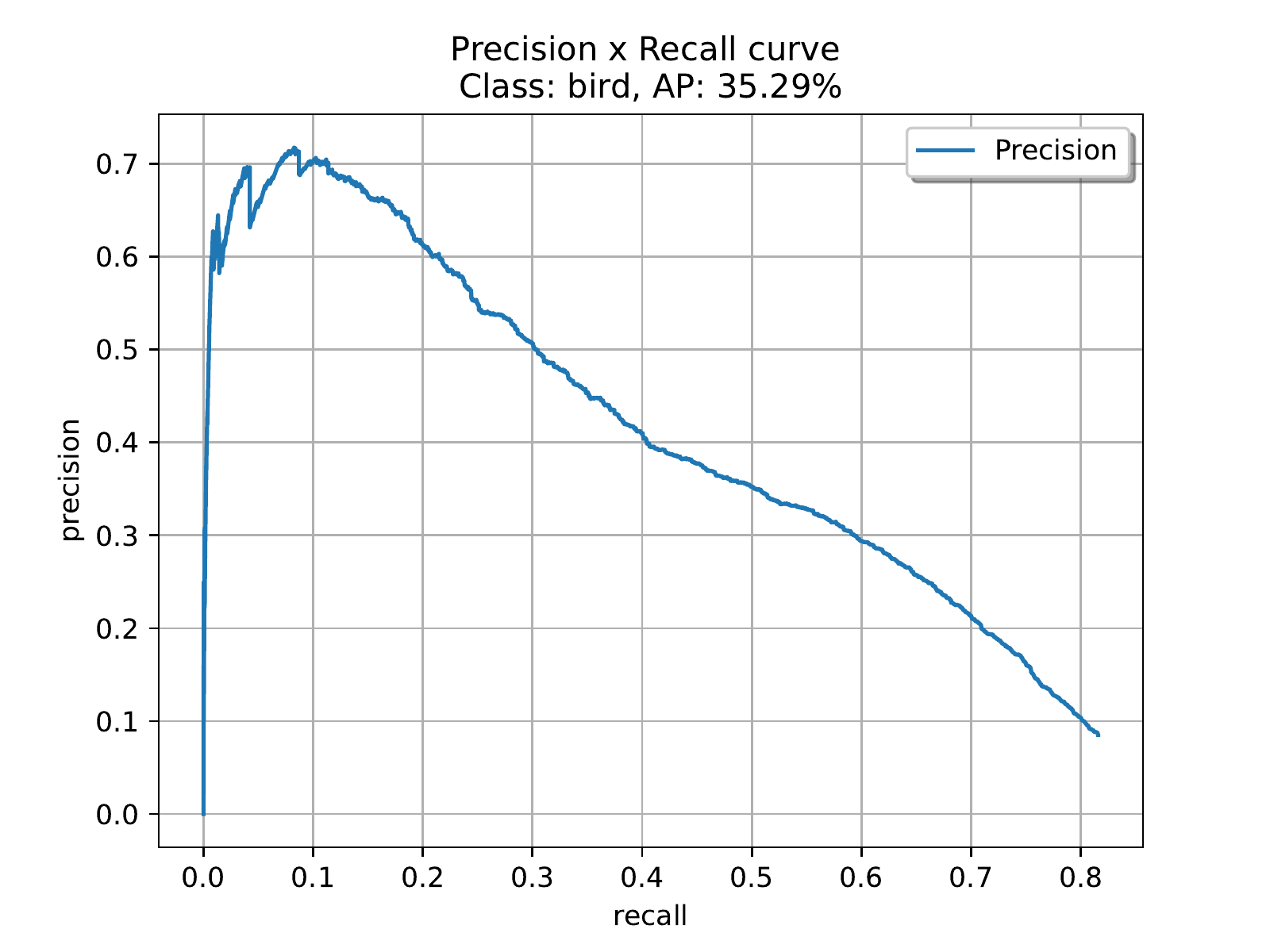}
    \caption{Faster}
    \label{fig:voc_pr_curve_faster}
  \end{subfigure}
  \begin{subfigure}{0.24\linewidth}
    \includegraphics[width=\linewidth]{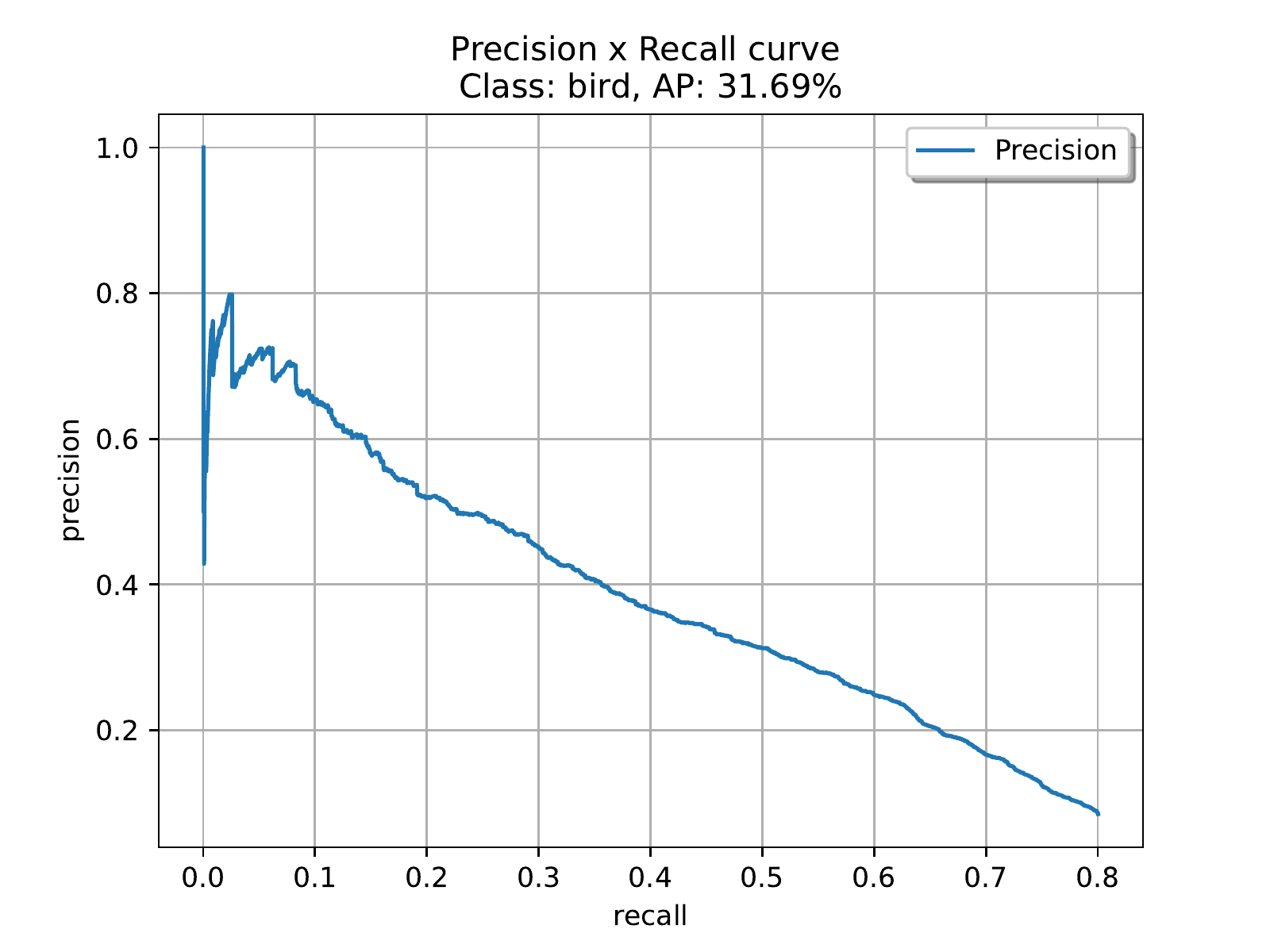}
    \caption{Cascade}
    \label{fig:voc_pr_curve_cascade}
  \end{subfigure}
  \begin{subfigure}{0.24\linewidth}
     \includegraphics[width=\linewidth]{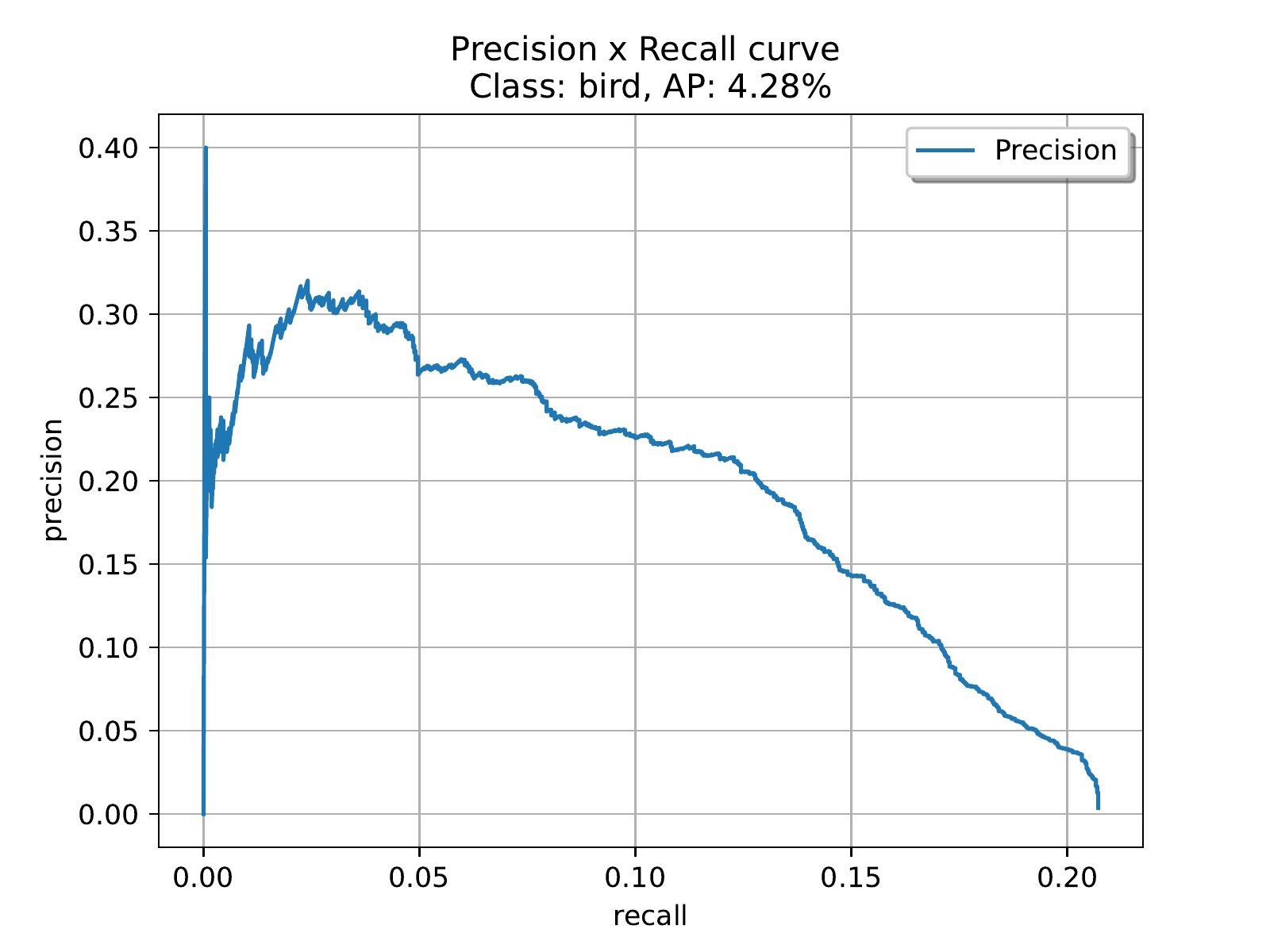}
     \caption{Deform}
     \label{fig:voc_pr_curve_deformable-detr}
  \end{subfigure}
  \begin{subfigure}{0.24\linewidth}
    \includegraphics[width=\linewidth]{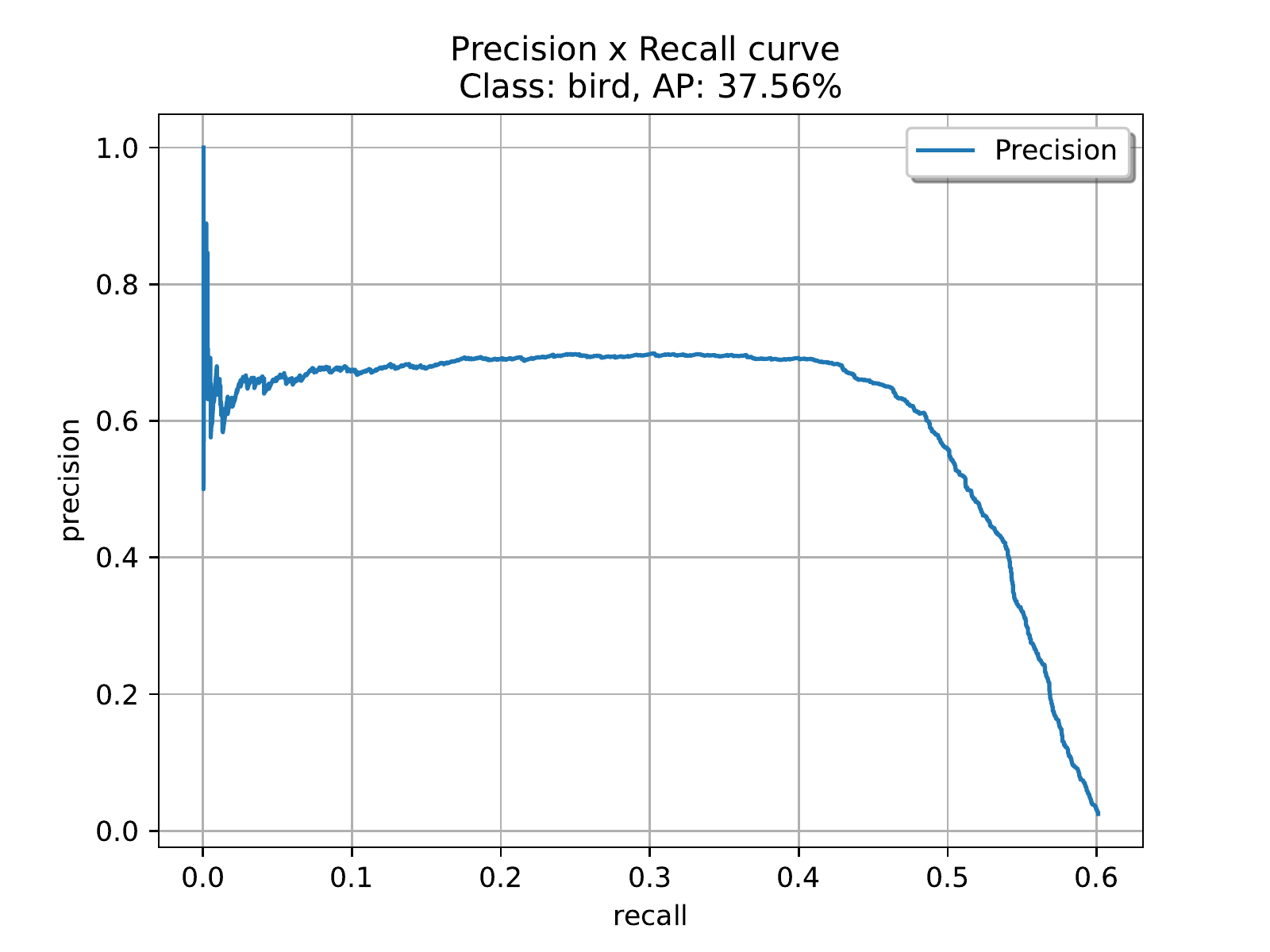}
    \caption{RepPoints}
    \label{fig:voc_pr_curve_reppoints}
  \end{subfigure}
  \begin{subfigure}{0.24\linewidth}
    \includegraphics[width=\linewidth]{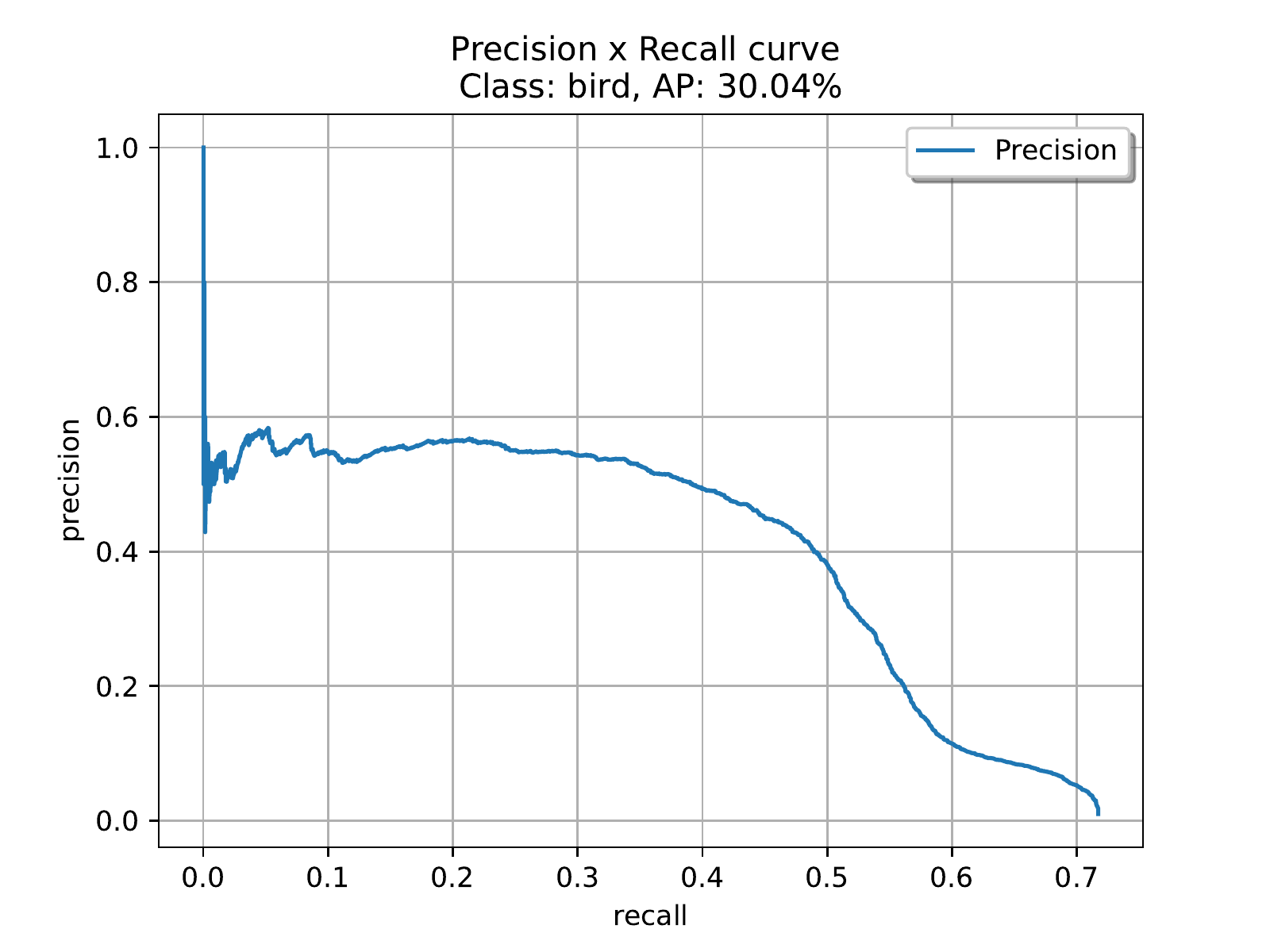}
    \caption{CornerNet}
    \label{fig:voc_pr_curve_cornernet}
  \end{subfigure}
  \begin{subfigure}{0.24\linewidth}
    \includegraphics[width=\linewidth]{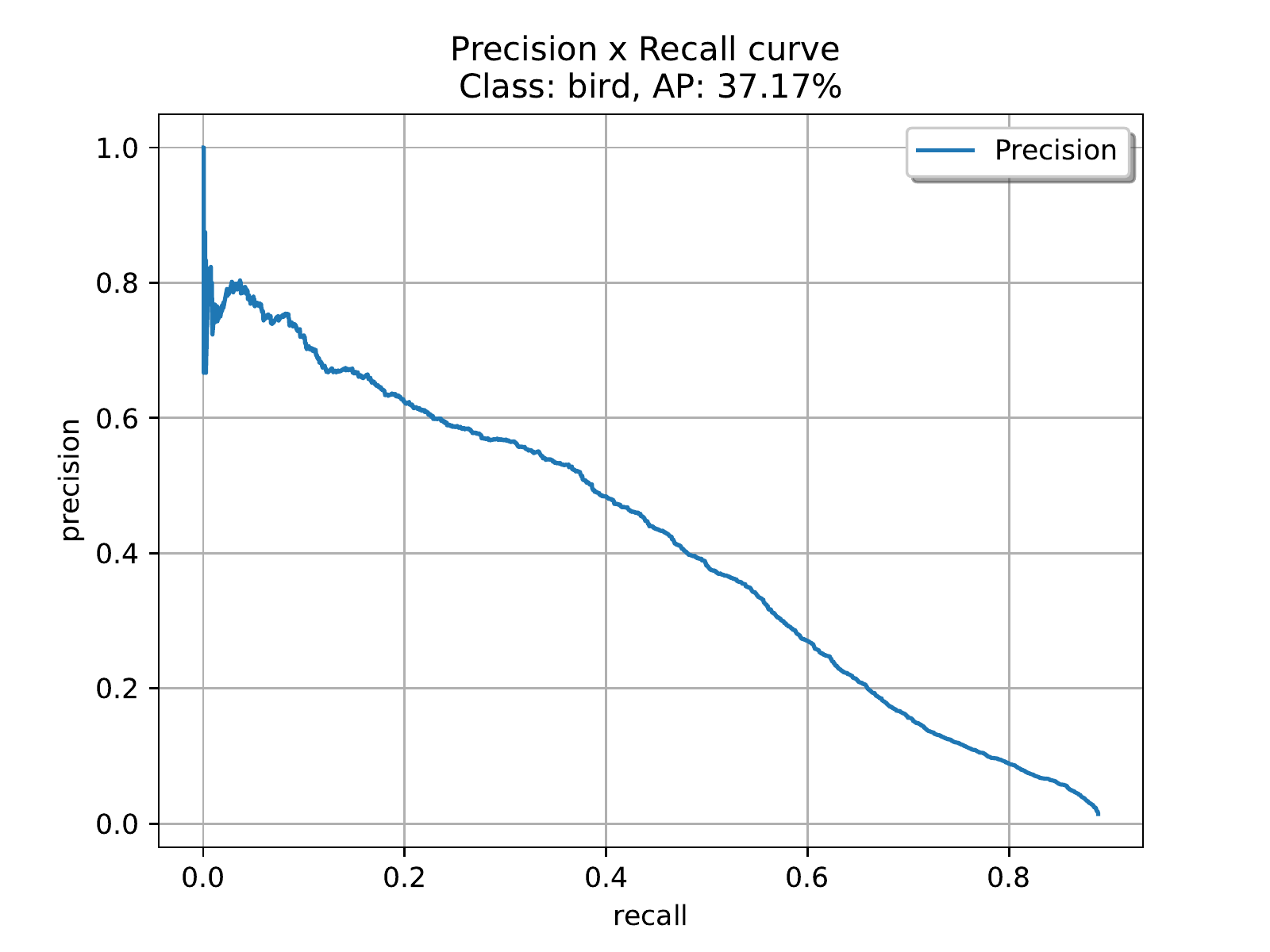}
    \caption{FreeAnchor}
    \label{fig:voc_pr_curve_freeanchor}
  \end{subfigure}
  \begin{subfigure}{0.24\linewidth}
    \includegraphics[width=\linewidth]{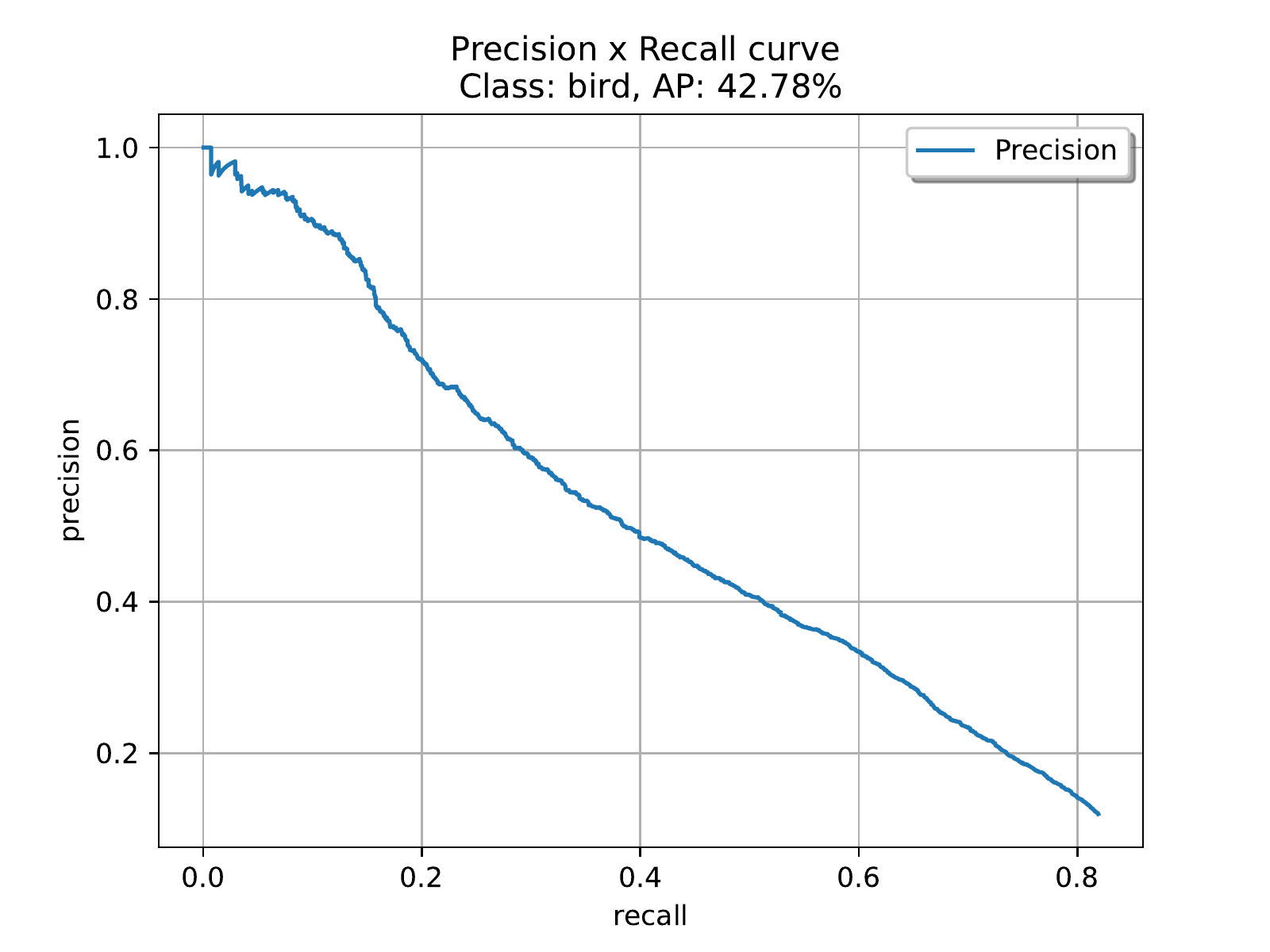}
    \caption{HRNet}
    \label{fig:voc_pr_curve_hrnet}
  \end{subfigure}
  \begin{subfigure}{0.24\linewidth}
    \includegraphics[width=\linewidth]{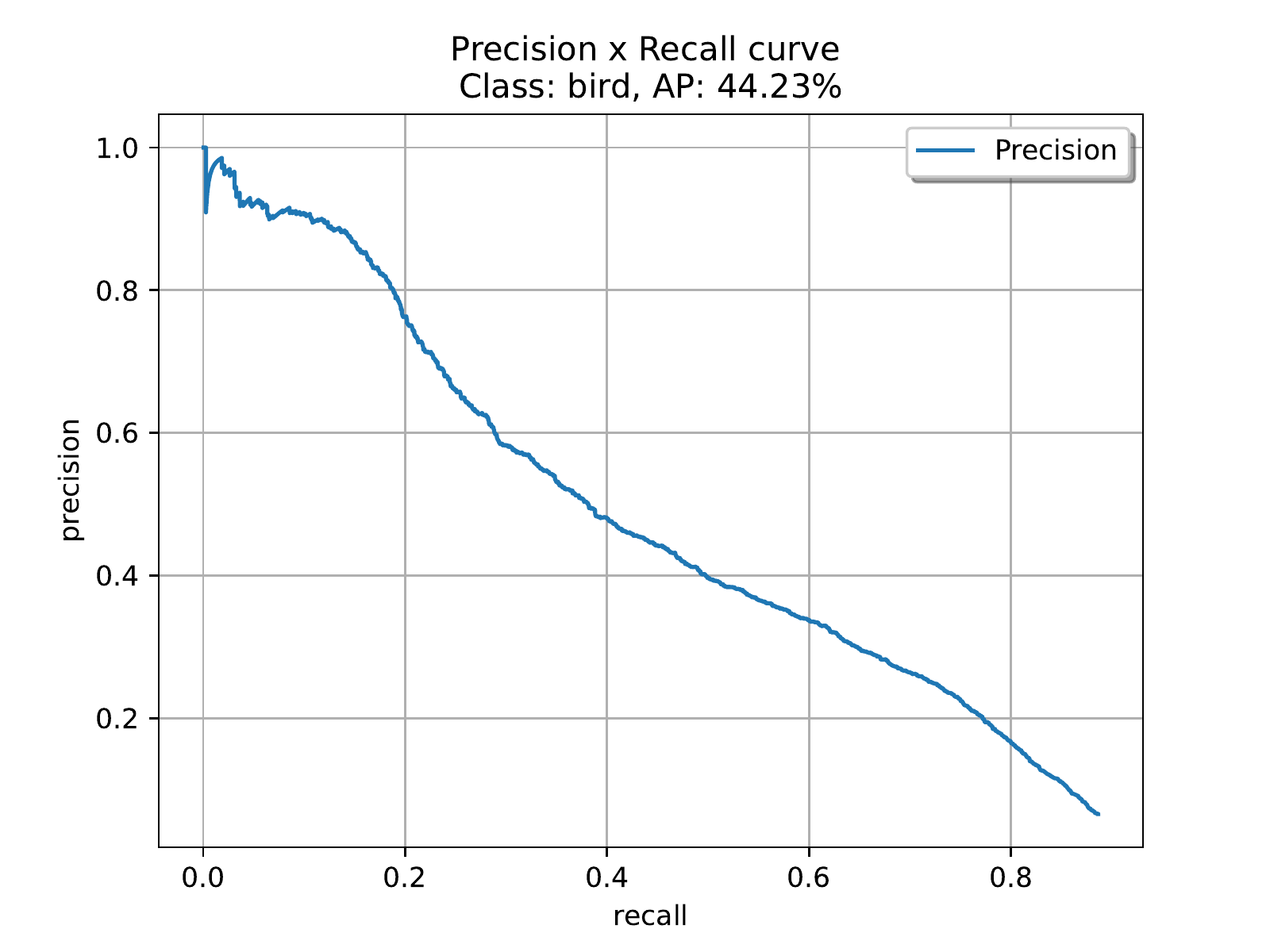}
    \caption{DCN}
    \label{fig:voc_pr_curve_dcn}
  \end{subfigure}
  \begin{subfigure}{0.24\linewidth}
    \includegraphics[width=\linewidth]{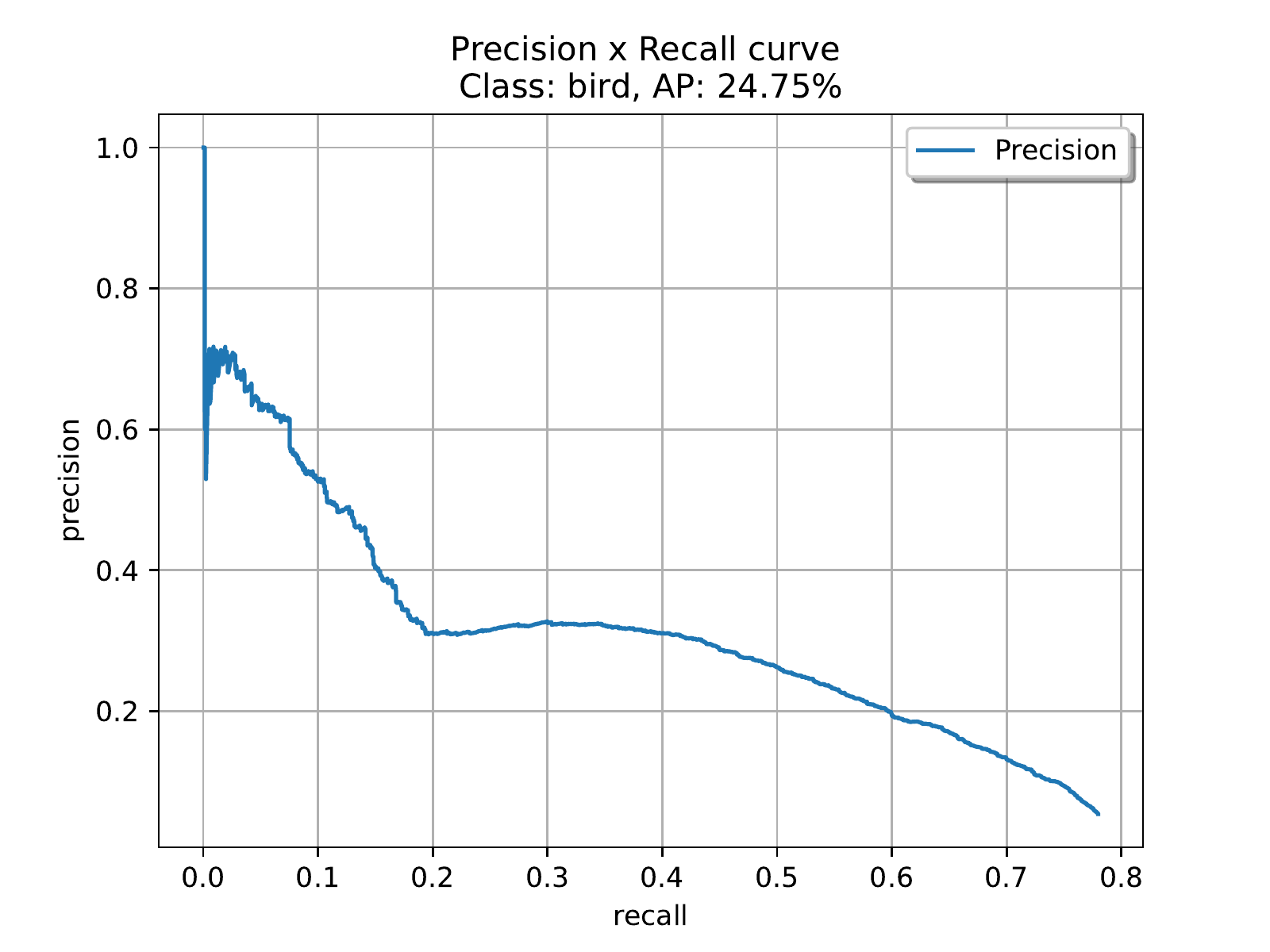}
    \caption{DCNv2}
    \label{fig:voc_pr_curve_dcnv2}
  \end{subfigure}
  \caption{Precision-Recall curves of different detectors in VOC\cite{Everingham10voc} style.}
  \label{fig:voc_pr_curve}
\end{figure}

\begin{figure}[!]
  \centering
  \begin{subfigure}{0.24\linewidth}
     \includegraphics[width=\linewidth]{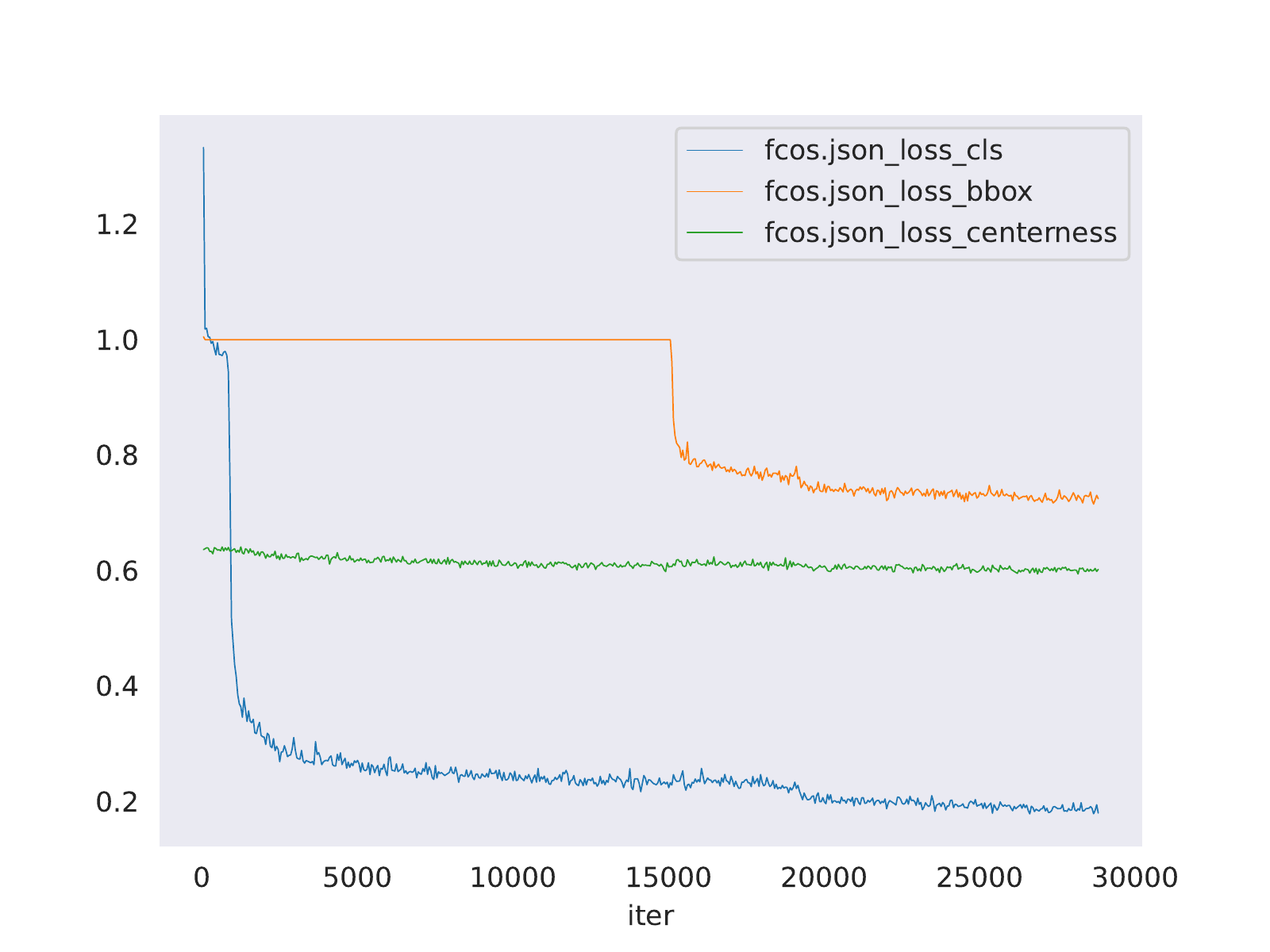}
     \caption{FCOS}
     \label{fig:fcos_loss}
  \end{subfigure}   
  \hfill
  \begin{subfigure}{0.24\linewidth}
     \includegraphics[width=\linewidth]{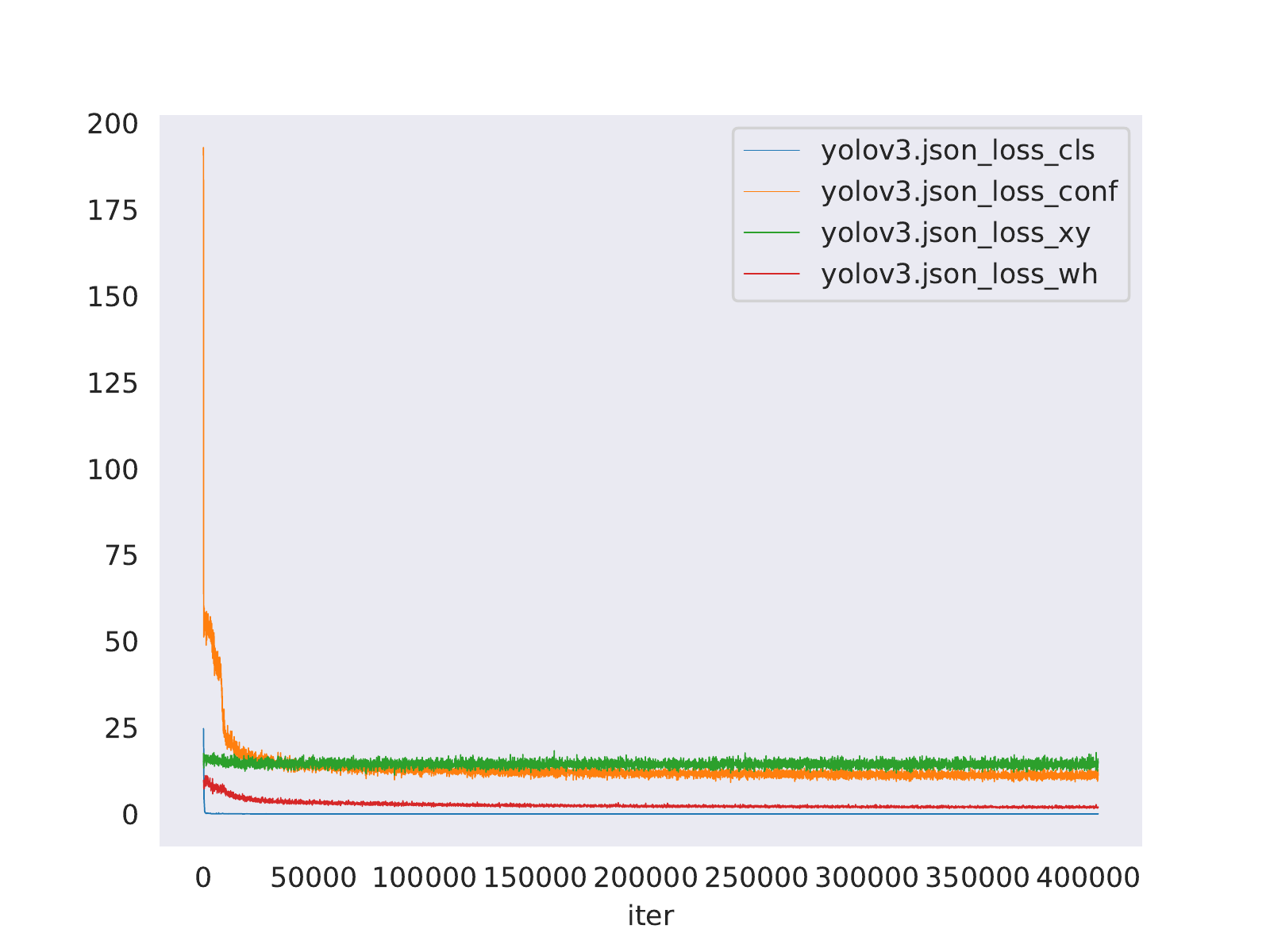}
     \caption{YOLOv3}
     \label{fig:yolov3_loss-detr}
  \end{subfigure}
  \hfill   
  \begin{subfigure}{0.24\linewidth}
     \includegraphics[width=\linewidth]{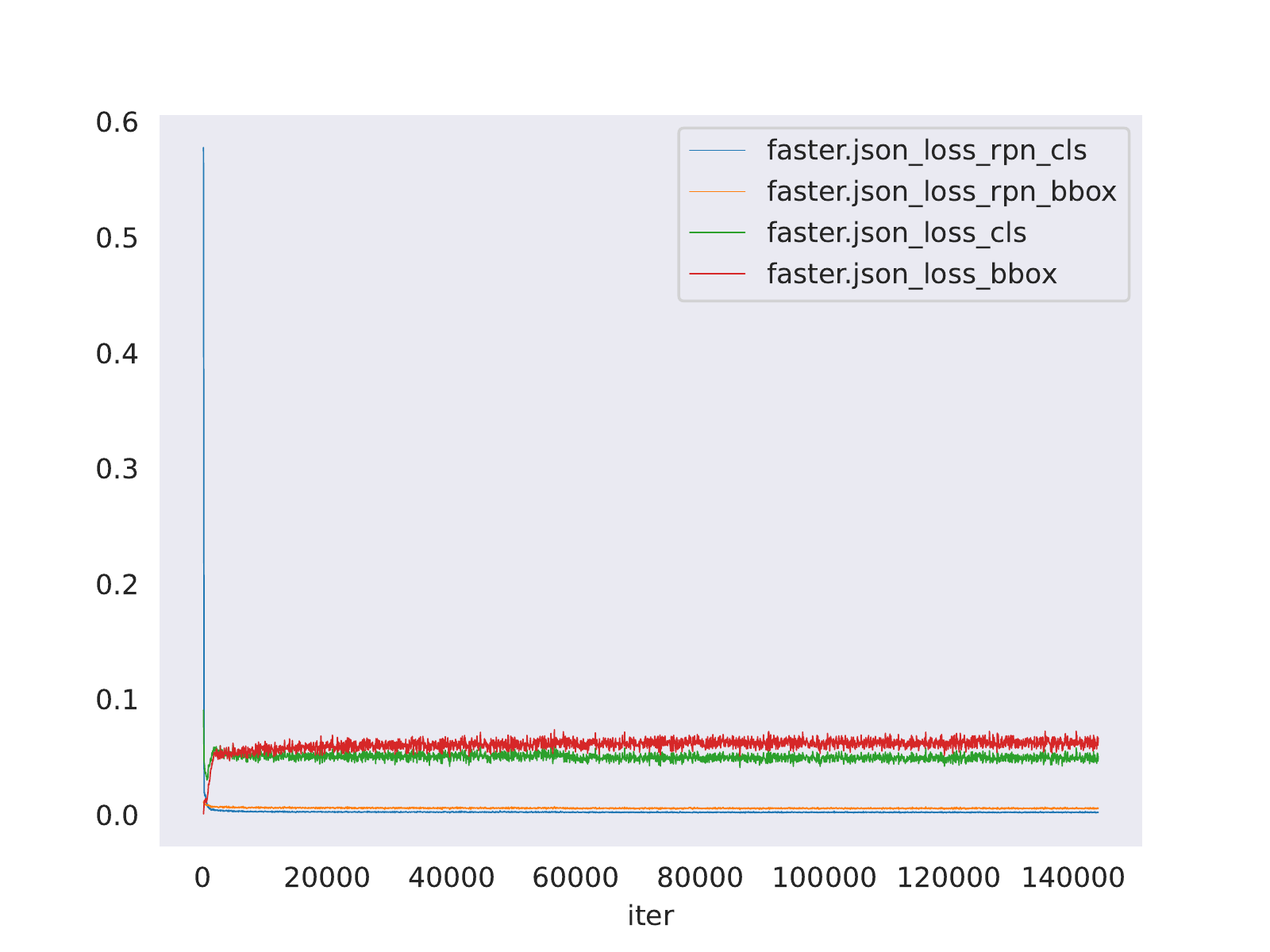}
     \caption{Faster}
     \label{fig:faster_loss}
  \end{subfigure}
  \hfill
  \begin{subfigure}{0.24\linewidth}
     \includegraphics[width=\linewidth]{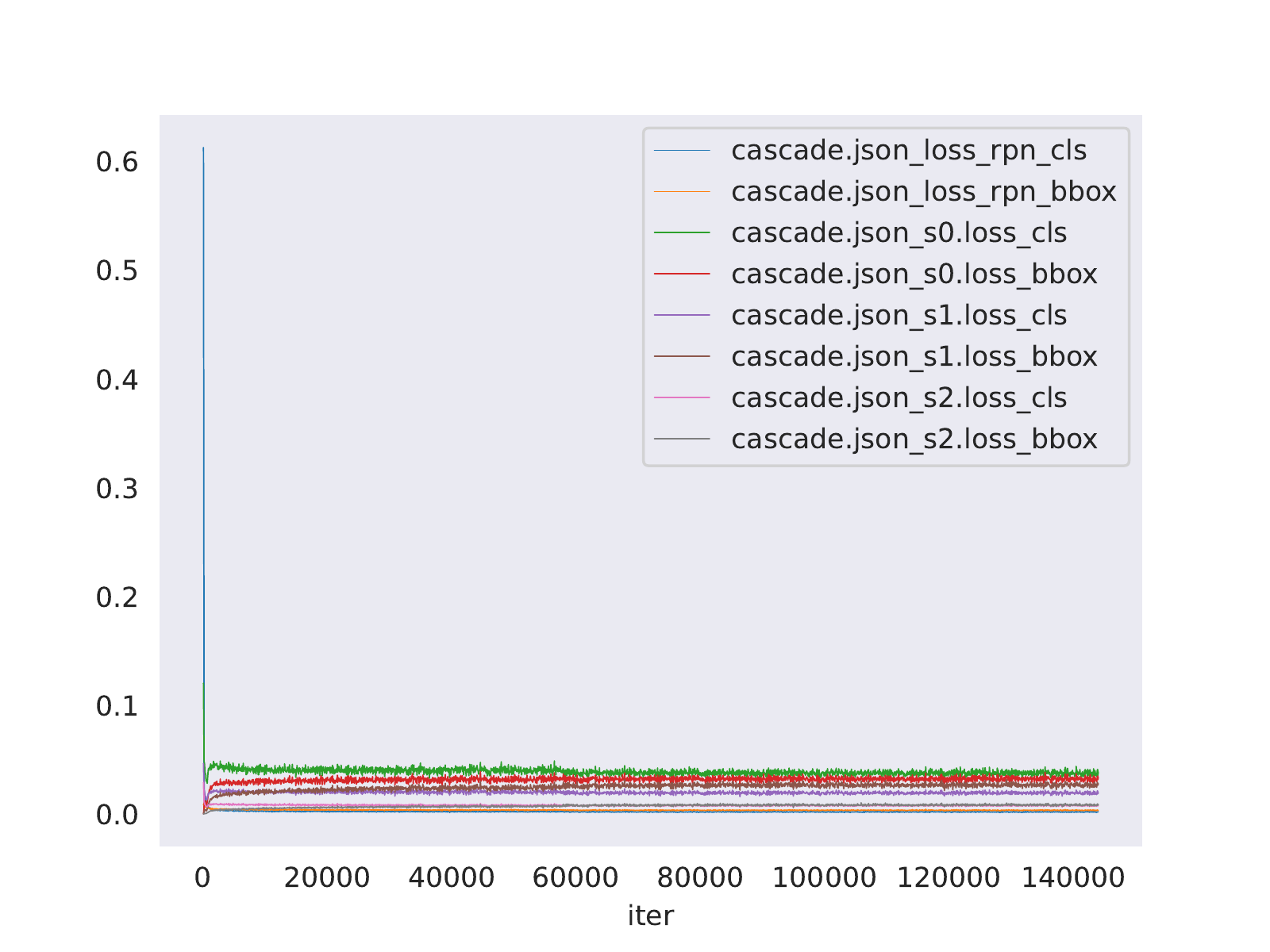}
     \caption{Cascade}
     \label{fig:cascade_loss}
  \end{subfigure}
  \begin{subfigure}{0.24\linewidth}
     \includegraphics[width=\linewidth]{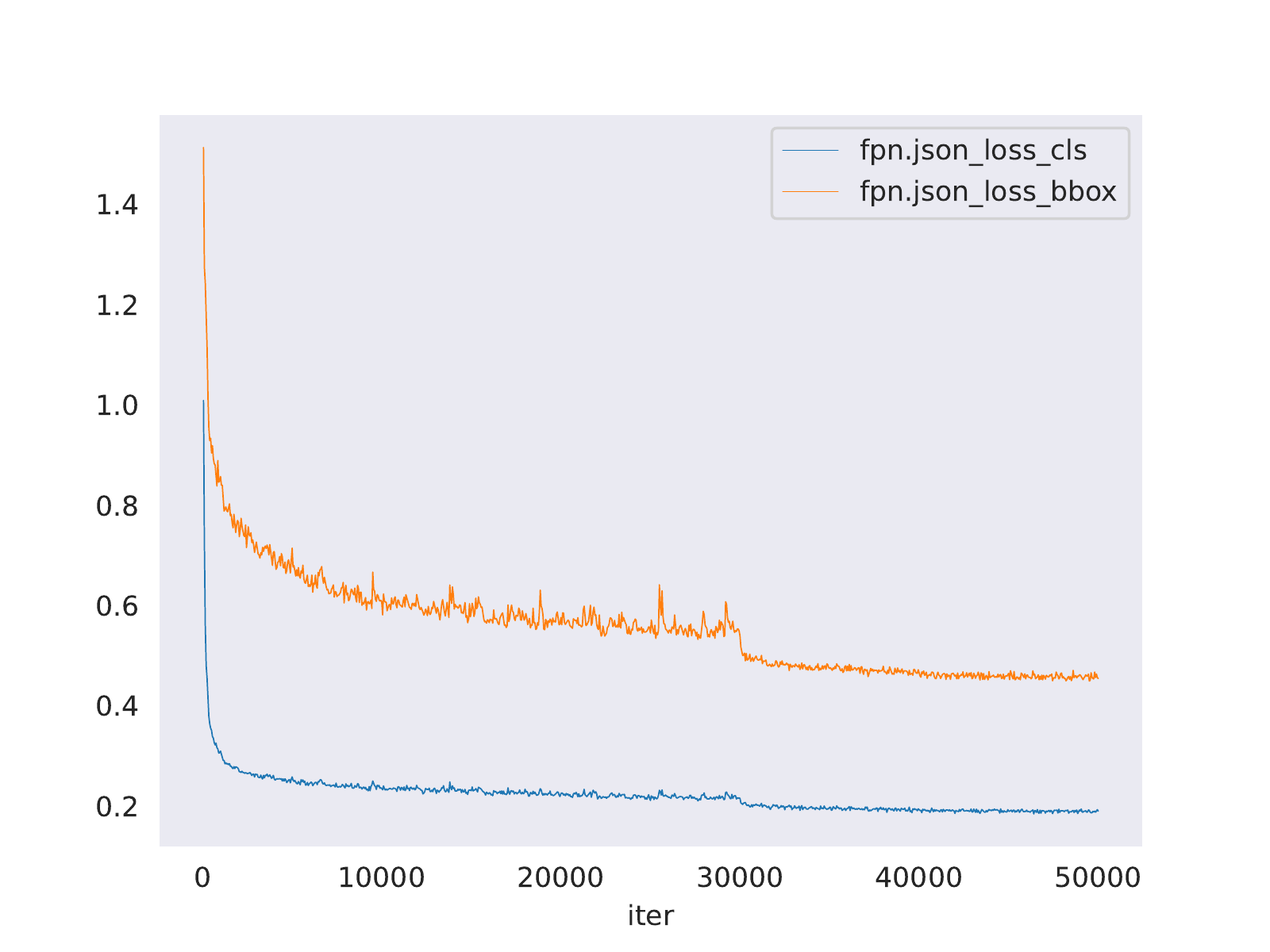}
     \caption{FPN}
     \label{fig:fpn_loss}
  \end{subfigure}
  \begin{subfigure}{0.24\linewidth}
    \includegraphics[width=\linewidth]{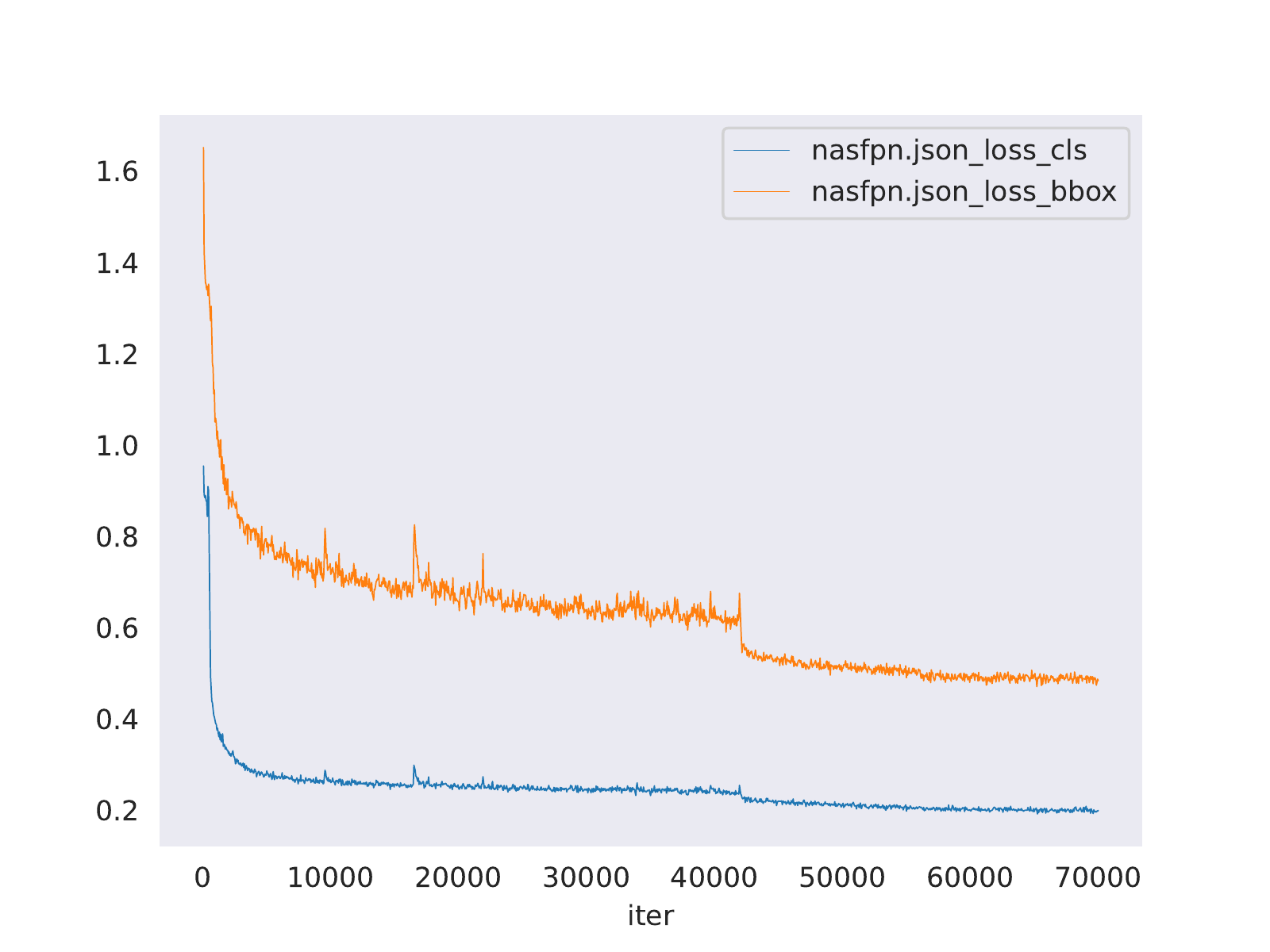}
    \caption{NASFPN}
    \label{fig:nasfpn_loss}
  \end{subfigure}
  \begin{subfigure}{0.24\linewidth}
    \includegraphics[width=\linewidth]{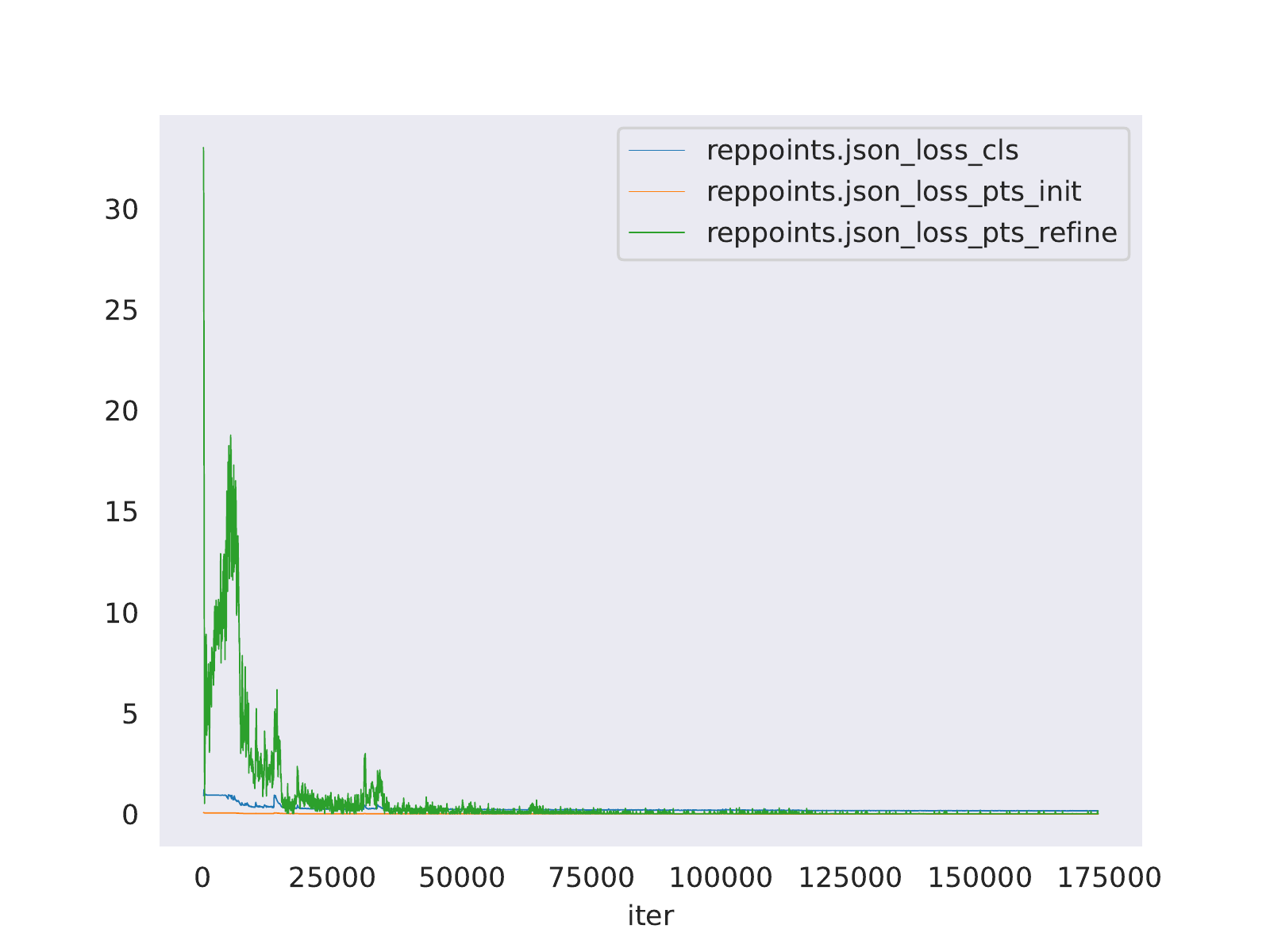}
    \caption{RepPoints}
  \label{fig:reppoints_loss}
  \end{subfigure}
  \begin{subfigure}{0.24\linewidth}
    \includegraphics[width=\linewidth]{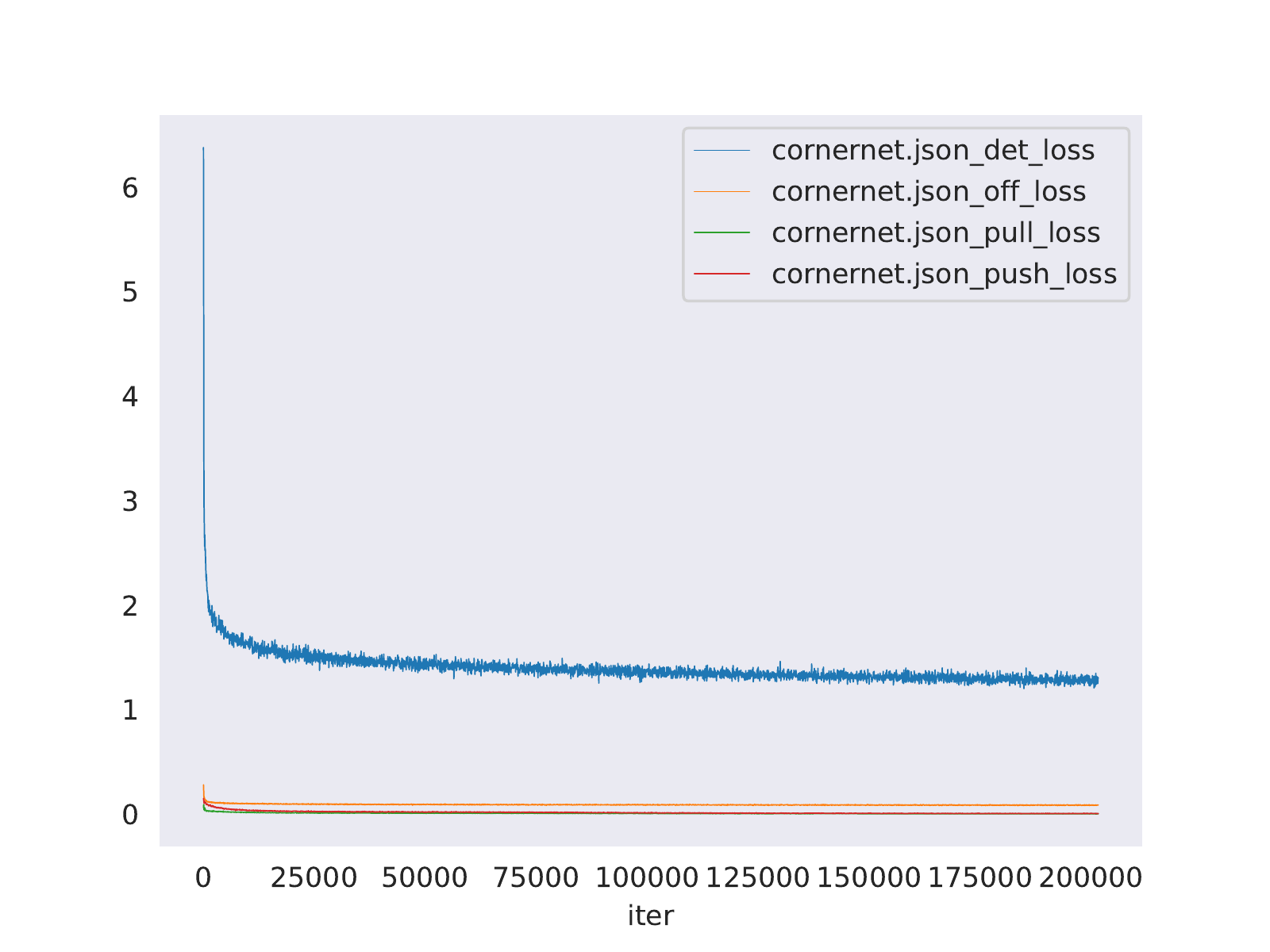}
    \caption{CornerNet}
    \label{fig:cornernet_loss}
  \end{subfigure}
  \begin{subfigure}{0.24\linewidth}
    \includegraphics[width=\linewidth]{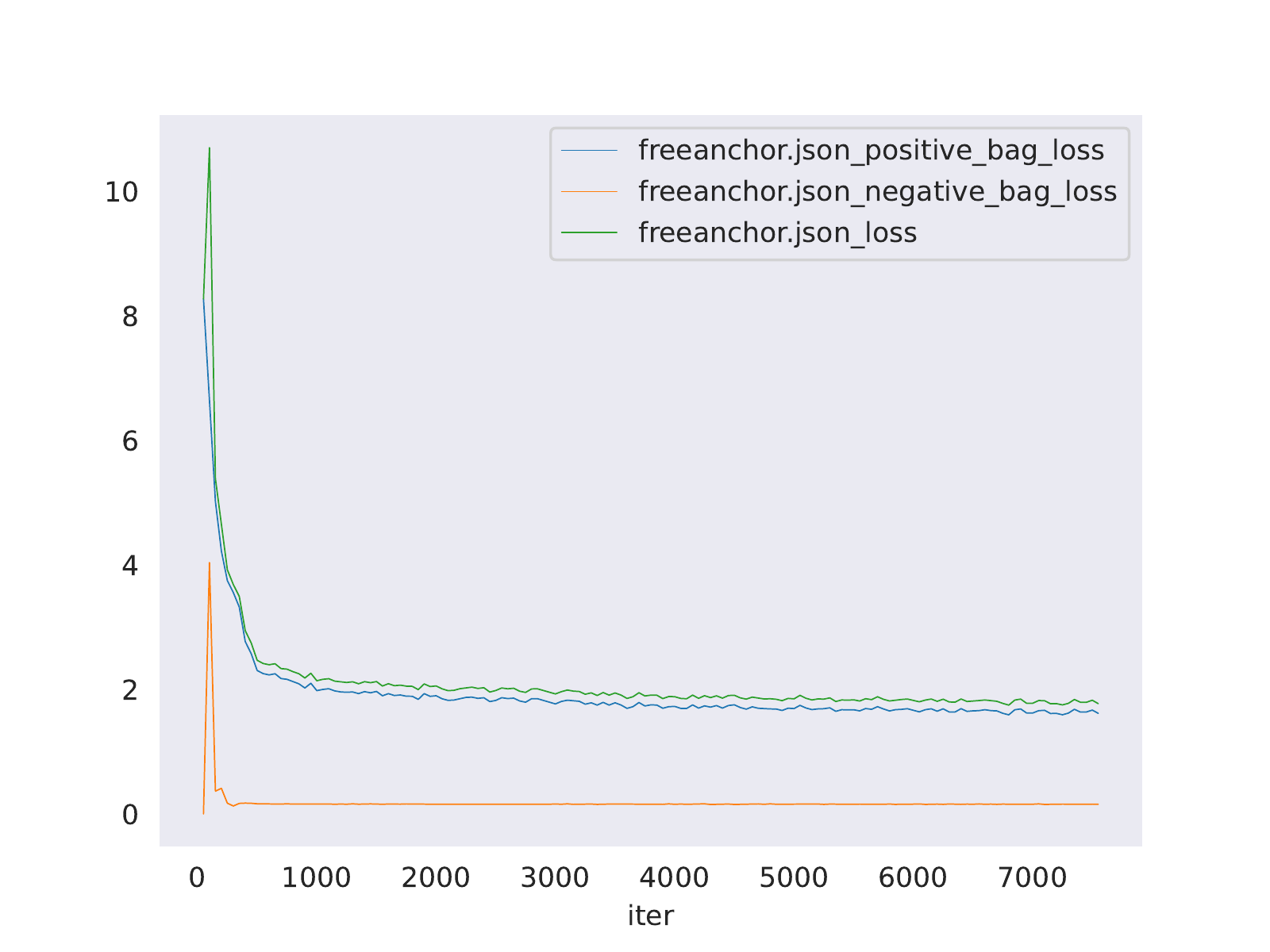}
    \caption{FreeAnchor}
  \label{fig:freeanchor_loss}
  \end{subfigure}
  \begin{subfigure}{0.24\linewidth}
    \includegraphics[width=\linewidth]{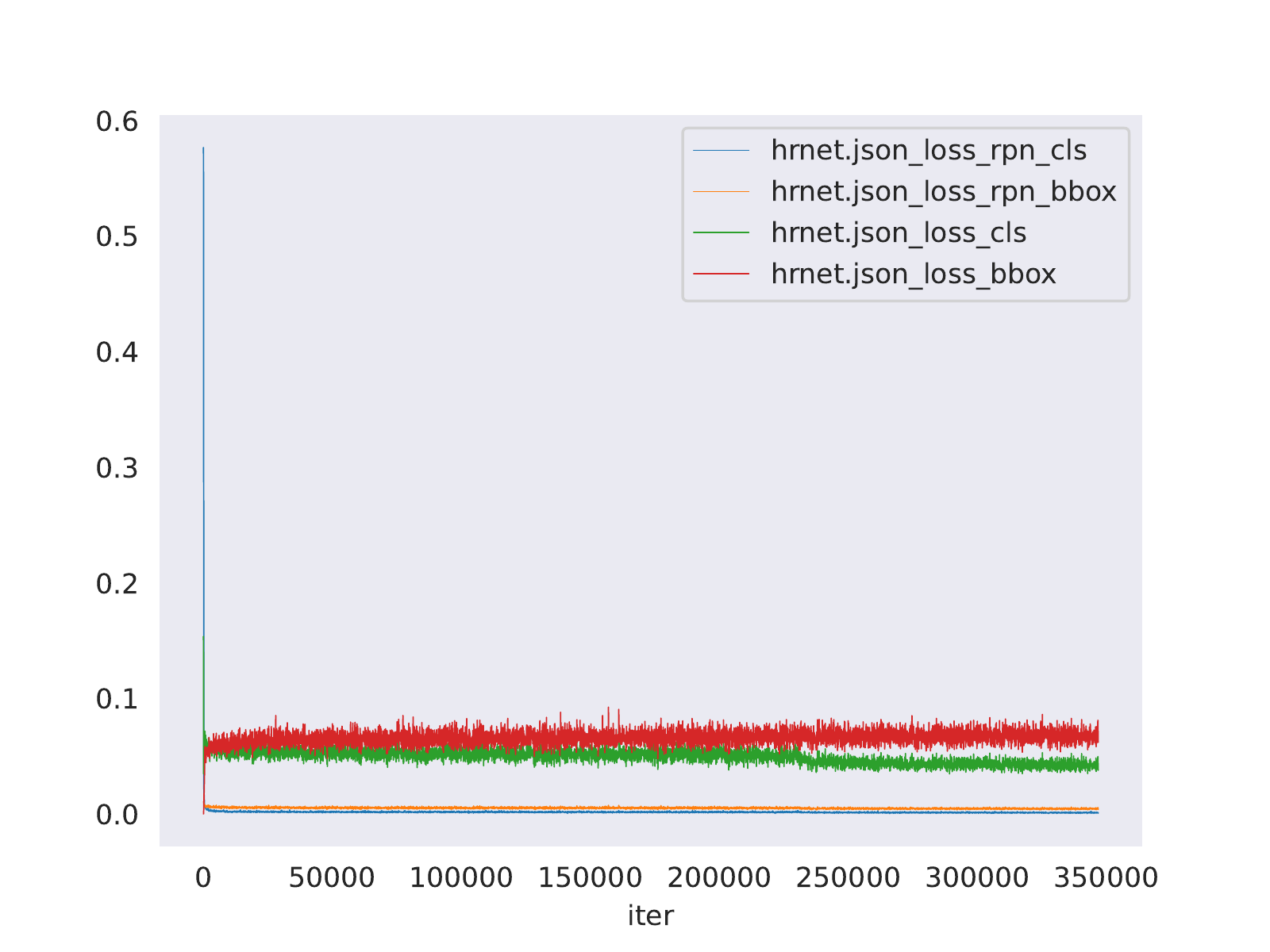}
    \caption{HRNet}
  \label{fig:hrnet_loss}
  \end{subfigure}
  \begin{subfigure}{0.24\linewidth}
    \includegraphics[width=\linewidth]{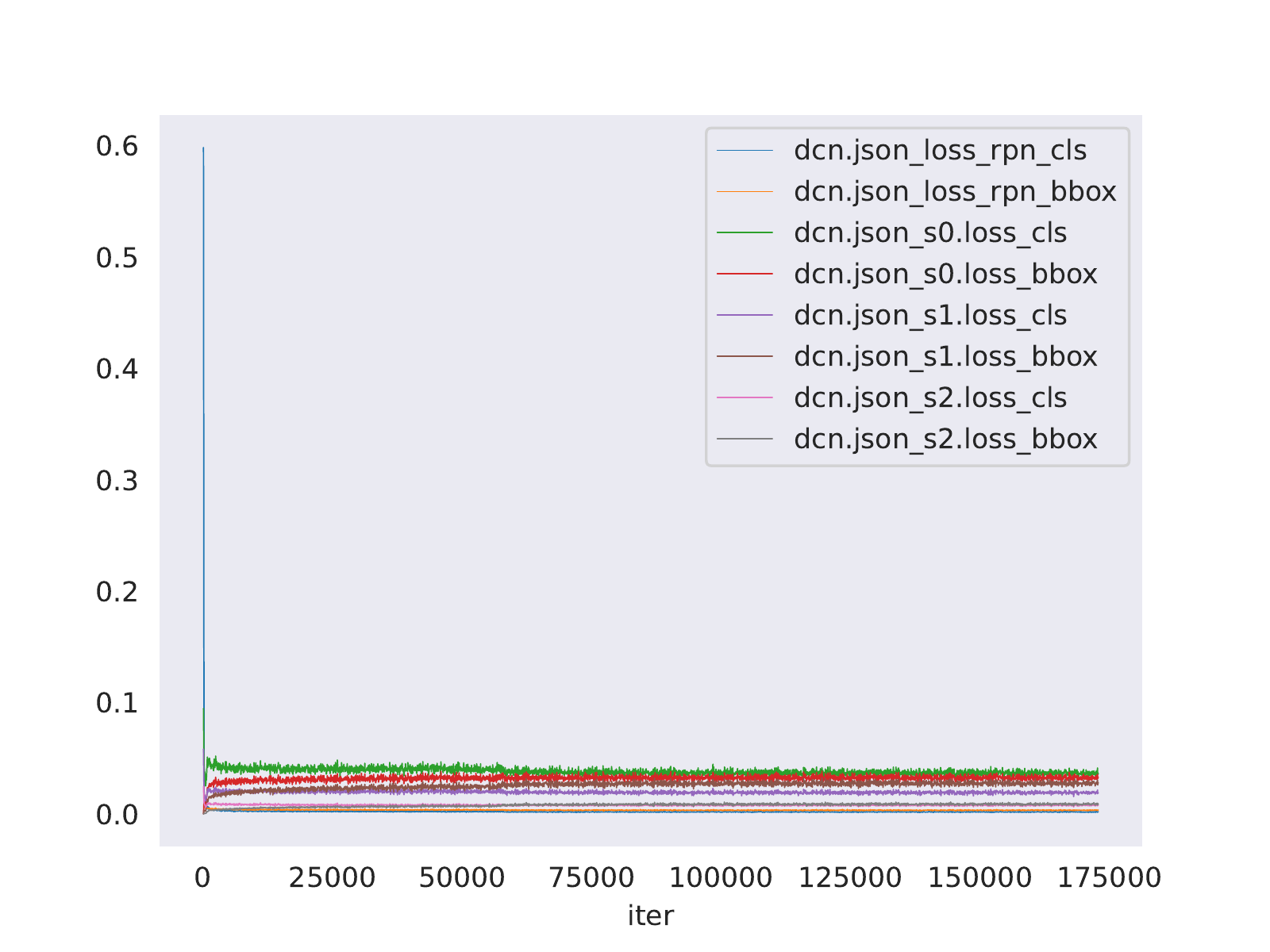}
    \caption{DCN}
  \label{fig:dcn_loss}
  \end{subfigure}
  \begin{subfigure}{0.24\linewidth}
    \includegraphics[width=\linewidth]{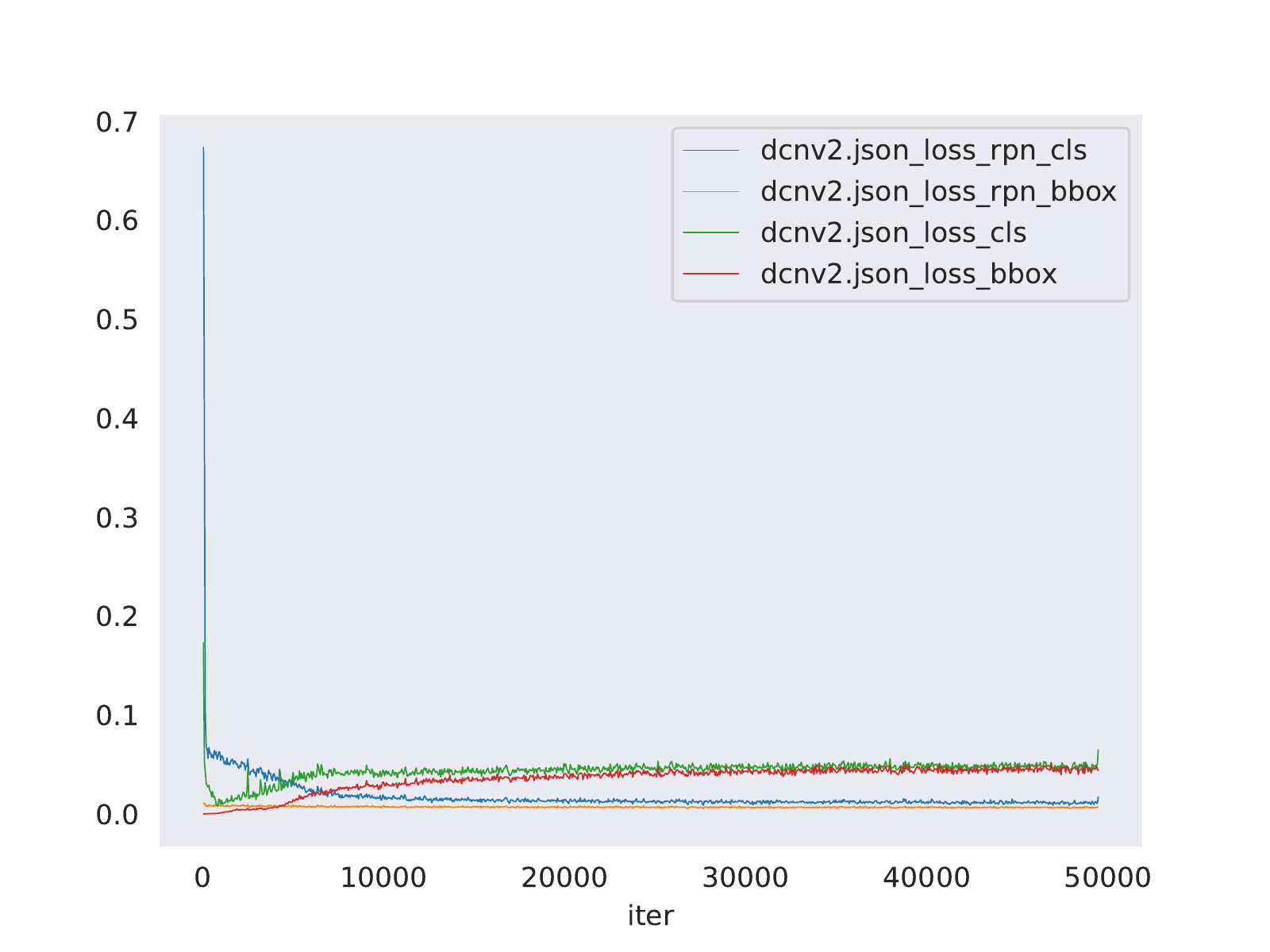}
    \caption{DCNv2}
  \label{fig:dcnv2_loss}
  \end{subfigure}  
  \caption{Training losses of different types of detectors.}
  \label{fig:detector_training_loss}
\end{figure}

\subsection{Results and Analysis}
Both accuracy and efficiency are equivalently important for bird discovering in a real-world airport. 
The accuracy is measured by $average~precision$ (AP) and the efficiency is judged by $frames~per~second$ (FPS).
Results are recorded in Tab.~\ref{tab:detection_results_on_airbirds}. 

For accuracy, the primary metric AP seems unsatisfactory, $e.g.$, the highest score achieved by
YOLOv5 is only 11.9, and the scores of all other models are less than 10.
We also compare the performances of those detectors on COCO and AirBirds, 
shown in Fig.~\ref{fig:map_coco_airbird}. 
Under the same detector, %
however, the performance gap is surprisingly large. For instance, 
the AP score of EfficientDet-D2 on COCO exceeds the one on AirBirds by 41.5(=42.1-0.6). 

Besides, precision-recall relationship are also investigated, 
and results are shown in Fig.~\ref{fig:voc_pr_curve}. The trend in all curves is that precision decreases
with increased recall because more and more false-positive birds produce as more and more birds are recalled.
YOLOv5 outperforms others while precision drops to 0 when recall reaches 0.7.

At this point, we wonder whether these detectors are well trained on AirBirds. 
Hence, their training losses are visualized in Fig.~\ref{fig:detector_training_loss}. 
We observe the losses of all detectors drop rapidly in the initial rounds of iterations, 
then progressively become smooth, indicating the training process is normal and converges to the target.

In terms of efficiency, YOLOv5 also outperforms others, surpassing 100 FPS on a 2080Ti GPU. 
However, most of detectors $fail~ to~ operate$ in \textbf{real-time efficiency} even with GPU acceleration,
which deviates a key principle of bird strike prevention. 

We also wonder why a wide range of detectors work poorly on AirBirds. Reasons are detailed 
in Section 3 in the supplementary material due to space limitation. 

In short, existing strong detectors show decent performances on commonly used datasets
$e.g.$ COCO, VOC etc. 
However, even with carefully customized configurations, they have room for significant improvements when validating on AirBirds.
The results also imply the non-trivial challenges of the research of bird strike 
prevention in real-world airports, where AirBirds can serve as a valuable benchmark.

\subsection{Effectiveness of the First Round of Annotations}
As mentioned in Section.~\ref{sec:airbirds_construction}, Alg.~\ref{alg:detection} provides the first round of bounding box
annotations for possible flying birds and the annotations are saved. 
Here we validate its effectiveness and compare it with the best performing YOLOv5. 
Different from $average~precision$ that sets strict IoU thresholds between detections and groundtruth,
actually precision, recall and f1 score are more meaningful metrics for evaluating initial annotations.

Table~\ref{tab:first_round_p_r_f1} shows Alg.~\ref{alg:detection} recalls more than 95\% of
birds in the initial round, which saves workers numerous efforts of 
discovering birds in subsequent rounds from scratch thus save costs.
In addition, the results indicate that sequence information is helpful for tiny flying birds detection
as the input images in Alg.~\ref{alg:detection} are in chronological order.
The star symbol in the second row in Tab.~\ref{tab:first_round_p_r_f1} means 
the results of Alg.~\ref{alg:detection} are obtained on an ordinary computer(i5 CPU, 16GB memory), without GPU support.

\section{Conclusion}
In this paper, we present AirBirds, a large-scale challenging dataset for bird strike prevention constructed directly from a real-world airport,
to close the notable gap of data distribution between real world and other tailor-made datasets.
Thorough statistical analysis and extensive experiments are conducted based on the developed dataset, revealing the non-trivial challenges
of bird discovering and bird strike prevention in real-world airports, which deserves increasing and further investigation,  
where AirBirds can serve as a first-hand and valuable benchmark.

We believe AirBirds will alleviate the fundamental limitation of the lack of a large-scale 
dataset dedicated for bird strike prevention in real-world airports, benefit researchers and  
the field. In the future, we will develop advanced detectors 
for flying bird discovering based on AirBirds. 

\subsubsection{Acknowledgements} 
We thank all members who involved in the system deploying, data collecting, processing and labeling. 
This work was supported in part by the National Natural Science Foundation of China (Grant No. 61972404, 12071478).

\bibliographystyle{splncs04}
\bibliography{egbib}

\end{document}